\documentclass[a4paper]{article}

\pdfoutput=1
\usepackage[utf8]{inputenc}

\usepackage{cite}

\usepackage{multirow}
\usepackage{longtable, caption}

\usepackage[english]{babel}
\usepackage[T1]{fontenc}
\usepackage{csquotes}
\usepackage{siunitx}

\usepackage[a4paper,top=3cm,bottom=2cm,left=3cm,right=3cm,marginparwidth=1.75cm]{geometry}

\usepackage{amsmath}
\usepackage{graphicx}
\usepackage{xr-hyper}
\usepackage{hyperref}

\makeatletter
\newcommand*{\addFileDependency}[1]{% argument=file name and extension
  \typeout{(#1)}
  \@addtofilelist{#1}
  \IfFileExists{#1}{}{\typeout{No file #1.}}
}
\makeatother

\hypersetup{colorlinks=true, allcolors=blue}
\usepackage{authblk}
\usepackage{multirow}
\usepackage{tabularx}
\usepackage{makecell}

\usepackage{float}
\usepackage[table,xcdraw]{xcolor}

\usepackage{listings}
\usepackage{color, colortbl}

\definecolor{Gray}{gray}{0.9}

\usepackage[font=small,labelfont=bf]{caption}
\usepackage{nomencl}
\usepackage{booktabs, array}
\newcolumntype{C}[1]{>{\centering\let\newline\\\arraybackslash\hspace{0pt}}m{#1}}

\usepackage{tabularx}
\usepackage{textcomp} 
\usepackage{amsfonts}
\usepackage{outlines} %%same
\usepackage{subfig}
\usepackage[version=4]{mhchem}
\usepackage{siunitx}
\usepackage{bm}
\usepackage[normalem]{ulem} %%temporary, for organization purposes
\usepackage{chemfig}
\usepackage{ifthen}

\date{}

\DeclareSIUnit\angstrom{\text {Å}}

\newfloat{sifigure}{htbp}{sif}
\floatname{sifigure}{Supplementary Fig.}

\floatstyle{plaintop}
\newfloat{sitable}{htbp}{sit}
\floatname{sitable}{Supplementary Table}
\usepackage[final, defaultcolor=purple]{changes}

\makeatletter
\newcommand\multiautoref[1]{\@first@ref#1,@}
\def\@throw@dot#1.#2@{#1}% discard everything after the dot
\def\@set@refname#1{%    % set \@refname to autoefname+s using \getrefbykeydefault
    \edef\@tmp{\getrefbykeydefault{#1}{anchor}{}}%
    \xdef\@tmp{\expandafter\@throw@dot\@tmp.@}%
    \ltx@IfUndefined{\@tmp autorefnameplural}%
         {\def\@refname{\@nameuse{\@tmp autorefname}s}}%
         {\def\@refname{\@nameuse{\@tmp autorefnameplural}}}%
}
\def\@first@ref#1,#2{%
  \ifx#2@\autoref{#1}\let\@nextref\@gobble% only one ref, revert to normal \autoref
  \else%
    \@set@refname{#1}%  set \@refname to autoref name
    \@refname~\ref{#1}% add autoefname and first reference
    \let\@nextref\@next@ref% push processing to \@next@ref
  \fi%
  \@nextref#2%
}
\def\@next@ref#1,#2{%
   \ifx#2@ and~\ref{#1}\let\@nextref\@gobble% at end: print and+\ref and stop
   \else, \ref{#1}% print  ,+\ref and continue
   \fi%
   \@nextref#2%
}
\makeatother

\author[1]{Thomas Marwitz}
\author[2]{Alexander Colsmann}  % ORCID 0000-0001-9221-9357
\author[3]{Ben Breitung}
\author[4,5]{Christoph Brabec}
\author[6]{Christoph Kirchlechner}
\author[7]{Eva Blasco}
\author[3]{Gabriel Cadilha Marques}
\author[3,8]{Horst Hahn}
\author[3,11]{Michael Hirtz}
\author[9]{Pavel A. Levkin}
\author[10]{Yolita M. Eggeler}
\author[3]{Tobias Schlöder}
\author[1,3,*]{Pascal Friederich}

\affil[1]{Institute of Theoretical Informatics, Karlsruhe Institute of Technology,\newline Kaiserstr.~12, 76131 Karlsruhe, Germany}
\affil[2]{Material Research Center for Energy Systems, Karlsruhe Institute of Technology,\newline Kaiserstr.~12, 76131 Karlsruhe, Germany}
\affil[3]{Institute of Nanotechnology, Karlsruhe Institute of Technology,\newline Kaiserstr.~12, 76131 Karlsruhe, Germany}
\affil[4]{Department of High Throughput Methods in Photovoltaics, Forschungszentrum Jülich GmbH, Helmholtz-Institute Erlangen-Nürnberg (HI ERN), Immerwahrstraße~2, 91058 Erlangen, Germany}
\affil[5]{Department of Materials Science and Engineering, Institute of Materials for Electronics and Energy Technology (i-MEET), Friedrich-Alexander-Universität Erlangen-Nürnberg, Martensstraße 7, 91058 Erlangen, Germany}
\affil[6]{Institute for Applied Materials,\newline Karlsruhe Institute of Technology, Kaiserstr.~12, 76131 Karlsruhe, Germany}
\affil[7]{Institute for Molecular Systems Engineering and Advanced Materials, Heidelberg University, Im Neuenheimer Feld 225, Heidelberg, 69120 Germany}
\affil[8]{Department of Materials Science and Engineering, University of Arizona, 1235 James E. Rogers Way, Tucson, AZ 85719, United States}
\affil[9]{Institute of Biological and Chemical Systems – Functional Molecular Systems, Karlsruhe Institute of Technology, Kaiserstr.~12, 76131 Karlsruhe, Germany}
\affil[10]{Laboratory for Electron Microscopy, Karlsruhe Institute of Technology,\newline Engesserstr.~7, 76131 Karlsruhe, Germany}
\affil[11]{Karlsruhe Nano Micro Facility, Karlsruhe Institute of Technology,\newline Hermann-von-Helmholtz-Platz 1, 76344 Eggenstein-Leopoldshafen, Germany}
\affil[*]{Corresponding author: pascal.friederich@kit.edu}

\title{Predicting New Research Directions in Materials Science using Large Language Models and Concept Graphs}

\begin{document}

%TC:ignore
\maketitle
\begin{abstract}
    Due to an exponential increase in published research articles, it is impossible for individual scientists to read all publications, even within their own research field.
    In this work, we investigate the use of large language models (LLMs) for the purpose of extracting the main concepts and semantic information from scientific abstracts in the domain of materials science to find links that were not noticed by humans and thus to suggest inspiring near/mid-term future research directions.
    We show that LLMs can extract concepts more efficiently than automated keyword extraction methods to build a concept graph as an abstraction of the scientific literature.
    A machine learning model is trained to predict emerging combinations of concepts, i.e. new research ideas, based on historical data.
    We demonstrate that integrating semantic concept information leads to an increased prediction performance.
    The applicability of our model is demonstrated in qualitative interviews with domain experts based on individualized model suggestions.
    We show that the model can inspire materials scientists in their creative thinking process by predicting innovative combinations of conceptsthat have not yet been investigated.
\end{abstract}
%TC:endignore

% Begin main text
Promising new research directions often arise from combining concepts that have previously not been investigated together \cite{uzzi_atypical_2013}. While experienced scientists possess vast domain knowledge enabling them to thoroughly explore research topics within (and adjacent to) their area(s) of expertise, finding new connections between their research topics and other yet unfamiliar topics to foster new ideas and findings is inherently challenging. Machine learning (ML) methods can help to look beyond the personal area of expertise by identifying previously unthought-of combinations of research topics, and thus enable the exploration of a vast hypothesis space beyond human intuition \cite{varshney_big_2013, pinel_culinary_2015}.\\
Scientific information is contained in a plethora of research publications in a rich but unstructured manner, and this lack of structured information poses challenges for automated analysis \cite{dunn_structured_2022, evans_advancing_2011}. Focusing on the extensive domain of material science, we first address the question of systematically extracting the main concepts of scientific articles, i.e. keywords or key phrases. Recent breakthroughs in Natural Language Processing (NLP) now allow the extraction of structured data from text and thereby enable the subsequent automated processing of this rich data \cite{vaswani_attention_2017, brown_language_2020, torfi_natural_2021, kojima_large_2023, bubeck_sparks_2023, norouzi_conexion_2025}. Here, we investigate whether large language models (LLMs) can offer improvements over traditional algorithmic methods in this extraction process.\\
After the identification and extraction of the concepts and their connections (i.e. the co-occurrence in the same article), we secondly investigate how to use this information to predict new combinations of concepts. In a prior study, Krenn et al. proposed SemNet, a graph that tracks the evolution of scientific literature in the domain of Quantum Physics \cite{krenn_predicting_2020}. The nodes of the SemNet graph are concepts, i.e. keywords extracted from text, using an algorithm called RAKE in conjunction with some predefined rules \cite{rose_automatic_2010}. Apart from analyzing emerging trends within SemNet, the authors exploit the evolution of the graph to predict its further development. To this end, they derive topological properties such as node degree and use them as input for a neural network (NN) to predict future connections. In a Kaggle challenge, the participants predicted changes in a SemNet built from AI literature \cite{krenn_predicting_2022}. While the most successful models combined specific hand-selected network features with machine learning techniques, such as NNs or Graph Neural Networks (GNNs), other participants employed purely theoretical or end-to-end ML approaches. All models of the participants could, however, only use the structure of SemNet as the real meaning behind the nodes was not revealed in the challenge. \\
In this study, the information on materials science concepts is similarly compressed into what we call a concept graph. Given the advances in language encoder models \cite{devlin_bert_2019, tenney_bert_2019}, we further use the MatSciBERT model \cite{gupta_matscibert_2021} to enrich the topological information of the nodes by additional information on the concepts in the form of semantic embeddings. After this, we further explore how the time evolution of this representation of the literature can be utilized by ML methods to perform link prediction.\\
Recent advances show that graph-based approaches can accelerate discovery in materials science: SciAgents employs multi-agent graph reasoning, Graph-PRefLexOR integrates symbolic graph abstractions with LLMs, and generative knowledge extraction with graph representations further supports hypothesis generation \cite{ghafarollahi_sciagents_2025, buehler_situ_nodate}. Complementary efforts on AI-driven ideation include SciMuse and SciMON, which use enriched co-occurrence and temporal knowledge graphs for novel idea generation, ResearchAgent, which iteratively refines literature-grounded ideas with knowledge-augmented LLMs, and SCI-IDEA, which applies context-aware embeddings for systematic ideation \cite{gu_interesting_2025, wang_scimon_2024, baek_researchagent_2025, keya_sci-idea_2025}.
In contrast to approaches which analyze understanding, intelligence, and creativity in general and try evoke these in machines \cite{schmidhuber_artificial_2010}, we hope to foster creativity in humans through the assistance of AI by providing material scientists with a tool to propose new research directions, i.e. previously uncombined concept combinations. To explore the real-world applicability of our model and its suggestions, we conduct interviews with researchers from the materials sciences. In these interviews, researchers assess how well suggested concepts generated by our model align with concepts from their own research.

\section*{Results}
\subsection*{Concept extraction and concept graph}

Using an LLM-based approach (see Methods), approximately 510,000 chemical formulae and 3,600,000 concepts were extracted from the 221,000 abstracts in our database, which corresponds to an average of 2.3 chemical formulae and 16.3 concepts per abstract. The extracted concepts were then condensed to approximately 52,000 unique formulae and 1,241,000 unique concepts by removing duplicates. In general, our method resulted in more precise concept extraction than rule-based approaches \ref{si-sec:llm-concept-evaluation}. Due to the extraction capabilities of LLMs, the amount of manual annotation work to generate the initially required data is negligible, especially as our iterative approach (see the Methods section below) reduces the manual effort to a minimum. Remarkably, the fine-tuned LLMs were able to extract concepts that were not exactly present in the text. \autoref{tab:llm-concept-extraction} shows selected examples to demonstrate the capabilities of fine-tuned LLMs for nominalisation, removal of fill words such as `of', plural-to-singular conversion, and formatting corrections. 

To construct a concept graph, we only include concepts that appeared at least 3 times and consist of at least 2 words. Due to this filtering, the resulting graph comprises approximately 137,000 nodes and 13,000,000 edges in total, making the calculation of topological features that require the squaring of the adjacency matrix possible in the first place. We present an overview of the 25 most frequently encountered concepts and formulas in \autoref{si-tab:frequent-concepts}.

An analysis of the node degree distribution in the resulting graph shows that the majority of the nodes have a degree between 30 and 1,000 (\autoref{si-fig:degree-distribution}). While a few concept nodes act as hubs -- having a notably larger number of connections than others -- most of the concepts in the graph are directly linked to only a few others, making the resulting graph sparse. The evolution of the concept graph over time shows that connectivity further increases, as more papers are being published using already existing concepts. We observe an increase in concept centralization, i.e. that fewer and fewer nodes account for a larger share of the total connections (\autoref{si-fig:concept-centralization}).

\begin{figure}
    \centering\small \includegraphics[width=\dimexpr1.0\textwidth\relax]{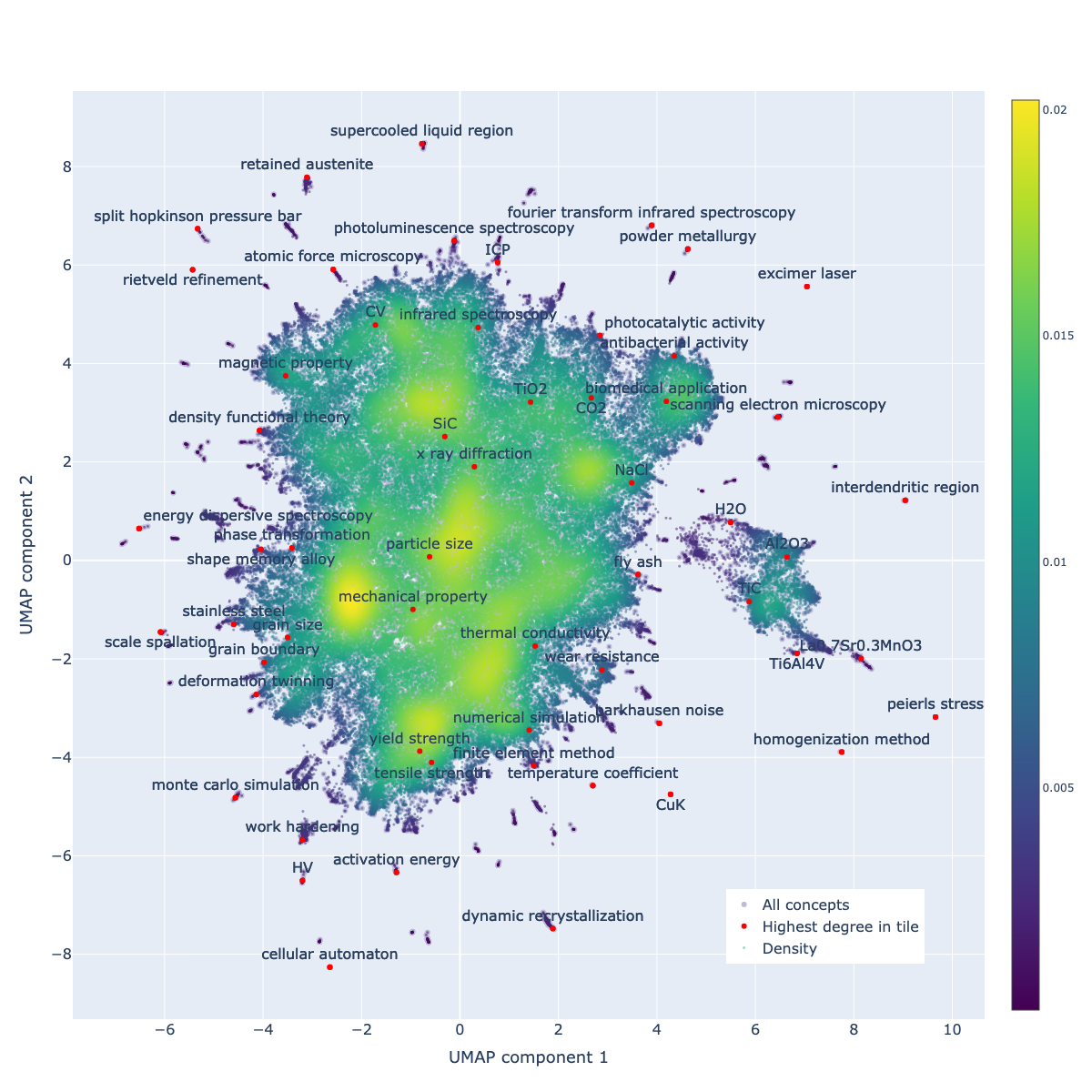}
    \caption{\textbf{Map of the materials science}. UMAP\cite{mcinnes_umap_2018} 2D projection of all extracted concepts with the highest-degree concepts in each square of length 2 highlighted and annotated (`Highest degree in tile'). Yellow and purple background colors respectively indicate high and low concept densities calculated using Kernel Density Estimation (KDE) is indicated by background color.}
    \label{fig:materials-science-map}
\end{figure}

We visualize all concepts by projecting their high-dimensional concept embeddings to 2 dimensions using UMAP \cite{mcinnes_umap_2018} with default settings. We display the result which we call the `Map of materials science' in \autoref{fig:materials-science-map} (an interactive version that can be explored on \href{http://inspire.aimat.science/}{inspire.aimat.science}). We then run nearest neighbor queries \cite{scikit-learn} on the concept embeddings to explore whether these 768-dimensional vectors capture semantic meaning. The example queries listed in \autoref{si-tab:knn} show the striking similarity between the queried concept and its nearest neighbors. 

\begin{table}
    \centering\small
    \caption{
        Selected examples of abstracts and concepts extracted by our fine-tuned Llama-2-13B model.
    }\begin{tabular}{p{0.4\textwidth}p{6pt}p{0.3\textwidth-12pt}p{6pt}p{0.3\textwidth}}\toprule
        Abstract excerpt && Extracted concept && Note\\\midrule
        "Strengthening mechanisms in short carbon fiber reinforced Nb/Nb5Si3 composites" && carbon fiber reinforcement && Nominalisation \\\midrule
        "Removal of Metal Impurities from the" && metal impurity removal && Removal of 'of', singular normalization \\\midrule
        "Resistance of Al2O3Coatings on Functional Structure" && al2o3 coating && Singular normalization, dealing with wrong formatting \\\midrule
        "Both fully and partially amorphous ribbons have been obtained" && fully amorphous ribbon, partially amorphous ribbon && Normalization of 'and' in concepts \\\midrule
         "Successive ionic layer adsorption and reaction SILAR trend for" && successive ionic layer adsorption and reaction && Extraction of long form without abbreviation in same concept \\\bottomrule
    \end{tabular}
    \label{tab:llm-concept-extraction}
\end{table}

\subsection*{Link prediction}

To statistically assess the performance of our link prediction models, we evaluate their performance on a held-out test set for edge formation in the period between 2020 and 2022, consisting of 2,000,000 node pairs and containing 307 (0.015\%) positives, i.e. emerging edges. We then complement this with a qualitative analysis of the real-world applicability of the models based on human expert knowledge.

\autoref{fig:roc-curves}~(a) shows the ROC curves for the prediction of link formation during the test period, as they illustrate the capacities of the model to distinguish between classes across all possible classification thresholds. ROC curves are particularly useful for imbalanced datasets because they evaluate performance independently of class distribution \cite{fawcett_roc_2004}. More information about test set creation, as well as detailed results (Precision/Recall@k) can be found in \autoref{si-sec:link-pred}.
Although the \emph{Baseline} model (a modified version of Krenn \emph{et al.} \cite{krenn_predicting_2022}; see Methods) performs slightly better (AUC 0.9109) than the \emph{Concept embeddings} (MatSciBERT) model (AUC 0.8855), the performance of the latter already shows that the semantic information contained in the concept embeddings can be used by our model architecture.  A \emph{GNN} model based on the GraphSAGE architecture (see Methods) surpasses the \emph{Baseline} with an AUC of 0.9288, suggesting that while both models access the same input features, the \emph{GNN} effectively leverages additional structural signals to improve performance. The \emph{Pure Text Baseline} (implemented via a fine-tuned MatSciBERT \cite{gupta_matscibert_2021}) also exploits this semantic information and performs similarly overall, though worse for  $d_{\text{prev}} = 3$ at a $5\times$ higher inference cost. Furthermore, the performance of the three hybrid models demonstrates that the link prediction task benefits from incorporating semantic knowledge on top of local graph features: While the \emph{Combination of features} model already shows a slightly improved AUC of 0.9147, the \emph{Mixture of Baseline and Embeddings} and \emph{Mixture of GNN and Embeddings} models approach exhibits a significant performance leap in the AUC metric, which reaches a maximum of 0.9433 when scaling the \emph{GNN} and \emph{Concept embeddings} model predictions by 0.5 and 0.5, respectively. 
We speculate that gradient descent optimization on a unified feature vector (concatenating the features of the \emph{Baseline} and \emph{Concept Embeddings} model) -- as it is done in the \emph{Combination of features} model -- might not be as effective as optimizing them individually. The distinct nature of baseline features versus high-dimensional concept embeddings could lead to the gradient for each batch becoming a suboptimal compromise between the gradients suited for each feature type in isolation.
While MatSciBERT may under-represent emerging or interdisciplinary concepts, it still benefits from the base BERT knowledge, and tokenization ensures meaningful embeddings. In our experiments, MatSciBERT (AUC 0.8855) outperformed BERT (AUC 0.8547), indicating an advantage of domain-specific embeddings, though BERT still offers a reasonable baseline.

\begin{figure}
    \centering\small
    \includegraphics[width=\textwidth]{img/fig2_roc-curves-all.png}
    \caption{Performance metrics (ROC and the respective AUC) for our link prediction models on the test set ($T_{\text{test}} = [2020, 2022]$). Markers highlight the performances at a threshold of $0.5$.
    (a) ROC curves on all data points with a zoomed-in view of the low false positive rate region in the inset. Panels (b) and (c) display the respective performance metrics for $d_{\text{prev}} = 2$ and $d_{\text{prev}} = 3$. Best result in bold.}
    \label{fig:roc-curves}
\end{figure}

We investigated the predicted new links with regard to the previous node distance $d_{\text{prev}}$, i.e. the shortest path distance between two nodes in $T_{\text{test}} = [2020, 2022]$ before they become directly connected. An analysis of the graph shows its dense interconnectedness as certain prevalent concepts in materials science, like `mechanical property’ and `x ray diffraction’ (\autoref{si-tab:frequent-concepts}) have edges with the majority of nodes, which leads to a large number of short distances between many concept pairs. Although the graph consists of 137,000 nodes, nearly all of the unlinked concept pairs in the test set were already connected through one ($d_{\text{prev}}=2$, 43.3\%) or two ($d_{\text{prev}}=3$, 56.5\%) concepts. The distribution of $d_{\text{prev}}$ is even more biased towards short paths for the positive samples, i.e. the connections that actually formed during the test period. 290 of 307 emerging edges (94.5\%) in the test set were found to have $d_{\text{prev}}=2$, while only 17 (5.5 \%) had a previous distance of 3 which clearly shows that the proximity of two nodes in the concept graph increases the probability for a new edge to form between them. Samples at $d_{\text{prev}}=4$ were found to be very scarce (0.2\% of the total samples, all negatives), and therefore excluded from further analysis.

While the \emph{Baseline} model tends to correctly predict emerging edges primarily at a distance of 2 (212/213 TP have $d_{\text{prev}}=2$) with a recall of 73.1\%, it performs much worse for $d_{\text{prev}}=3$ (recall: 5.9\%). By contrast, the \emph{Concept embeddings} model achieves a significantly better recall of 35.3\% (p < 0.05, DeLong test \cite{delong_comparing_1988} for $d_{\text{prev}}=3$ while only slightly compromising on the recall at $d_{\text{prev}}=2$ (70.0\%). Interestingly, the \emph{GNN} model matches this performance at $d_{\text{prev}}=3$, demonstrating that improved structural processing can rival the benefits of semantic embeddings for distant connections. The results are summarized in \ref{tab:ed-tab-confusion}. The high number of false positives, especially at $d_{\text{prev}}=3$ is not a problem in itself, as those combinations may remain scientifically plausible and will subsequently be evaluated by human scientists. Hence, we prioritize recall over precision in order not to miss valuable ideas.

\begin{table}[h!]
    \centering\small
    \caption{Confusion matrix classes for the link predictions of the baseline, Concept Embeddings, GNN, and GNN-Mixture models split by previous node distances.}
    \label{tab:ed-tab-confusion}
    \begin{tabular}{C{0.125\textwidth}C{0.075\textwidth}C{0.1\textwidth}C{0.1\textwidth}C{0.1\textwidth}C{0.1\textwidth}C{0.1\textwidth}}\toprule
         & & \multicolumn{2}{c}{Positives} & & \multicolumn{2}{c}{Negatives}\\ \cmidrule(lr){3-4}\cmidrule(lr){6-7}
         Model & $d_{\text{prev}}$ & TP & FN & Recall & FP & TN\\ \midrule
         \multirow{3}{*}{\textit{Baseline}} & 2 & 212 & 78 & 73.1\% & 115,807 & \phantom{0,}751,021\\ 
         & 3 & \phantom{00}1\ & 16 & \phantom{0}5.9\% & \phantom{00}5,342 & 1,124,056\\
         & 4 & \phantom{00}0 & \phantom{0}0 & \multicolumn{1}{c}{--} & \phantom{000,00}0 & \phantom{0,00}3,467 \\ \midrule
         \multirow{3}{*}{\textit{\shortstack{Concept\\embeddings}}} & 2 & 203 & 87 & 70.0\% & 120,575 & \phantom{0,}746,253 \\
         & 3 & \phantom{00}6 & 11 & \phantom{0}35.3\% & \phantom{0}26,209 & 1,103,189\\
         & 4 & \phantom{00}0 & \phantom{0}0 & \multicolumn{1}{c}{--} & \phantom{000,0}27 & \phantom{0,00}3,440 \\ \midrule
         \multirow{3}{*}{\textit{GNN}} & 2 & 253 & 37 & 87.2\% & 223,637 & \phantom{0,}643,191 \\
         & 3 & \phantom{00}3 & 14 & \phantom{0}17.6\% & \phantom{0}31,512 & 1,097,886 \\
         & 4 & \phantom{00}0 & \phantom{0}0 & \multicolumn{1}{c}{--} & \phantom{000,00}9 & \phantom{0,00}3,458 \\ \midrule
         \multirow{3}{*}{\textit{\shortstack{Mixture of \\ GNN and \\Embeddings}}} & 2 & 248 & 42 & 85.5\% & 153,455 & \phantom{0,}713,373 \\
         & 3 & \phantom{00}5 & 12 & \phantom{0}29.4\% & \phantom{0}11,753 & 1,117,645 \\
         & 4 & \phantom{00}0 & \phantom{0}0 & \multicolumn{1}{c}{--} & \phantom{000,00}6 & \phantom{0,00}3,461 \\ \bottomrule
    \end{tabular}
\end{table}
Note that optimizing the classification metrics by changing the classification threshold of 0.5 for a positive prediction is outside the scope of this work, which mainly aims at rating non-existing links with respect to their future emergence rather than accurately predicting whether a new link will form or not.

In addition, we also separately calculated the ROC curves and the corresponding AUCs for both $d_{\text{prev}}=2$ and $d_{\text{prev}}=3$, and the results (\autoref{fig:roc-curves} (b) and (c)) emphasize the \emph{Baseline} model’s failure to correctly categorize most positives with $d_{\text{prev}}=3$. This not only highlights the inherent challenge of predicting positives at greater distances but also indicates that the integration of semantic information enhances the model’s ability to forecast connections between concept pairs that are further apart in the graph. However, these emerging connections with larger previous node distances are particularly interesting and hold great potential to broaden the scientific scope beyond the more obvious new research directions. Ultimately, the \emph{Mixture of GNN and Embeddings} yields the highest AUC for these distant connections, outperforming the individual models by effectively combining structural and semantic signals.

We evaluated our Baseline model on the Science4Cast benchmark, where it achieved an AUROC of 0.9088, ranking second among all reported approaches in \cite{krenn_predicting_2022}. This demonstrates that a deep neural network trained on a large set of semantically meaningful features can outperform most competing methods, including those based on Common Neighbors and node2vec embeddings combined with a Transformer architecture \cite{vaswani_attention_2017, liben-nowell_linkprediction_2007, grover_node2vec_2016}. As Science4Cast does not contain any semantic or text information about the meaning of nodes, we cannot apply and benchmark our embedding-based models on the Science4Cast challenge.

\subsection*{Human expert evaluation: Analysis} 
\label{sec:interviews}

As the second part of the model performance analysis, we furthermore conducted interviews with ten material scientists (human experts) based on individualized reports containing individualized recommendations for combinations of concepts suggested by our prediction model (\emph{Mixture of Baseline and Embeddings}). Personalized recommendations were generated using the \emph{Mixture of Baseline and Embeddings} model which is marginally less performant ($<0.01$ AUC) than the \emph{Mixture of GNN and Embeddings} model, because the GNNs were only tested at a later stage of our study. The suggestions were subsequently discussed in the interviews to assess and clarify the proposed concept combinations. The small sample size, i.e. the small number of interviewees, as well as a potential selection bias limit the robustness of our conclusions and only allow a qualitative analysis and anecdotal findings. Nonetheless, the expert feedback sheds light on the usefulness of the suggestions provided by our model.

\begin{figure}
    \centering\small
    \includegraphics[width=\textwidth]{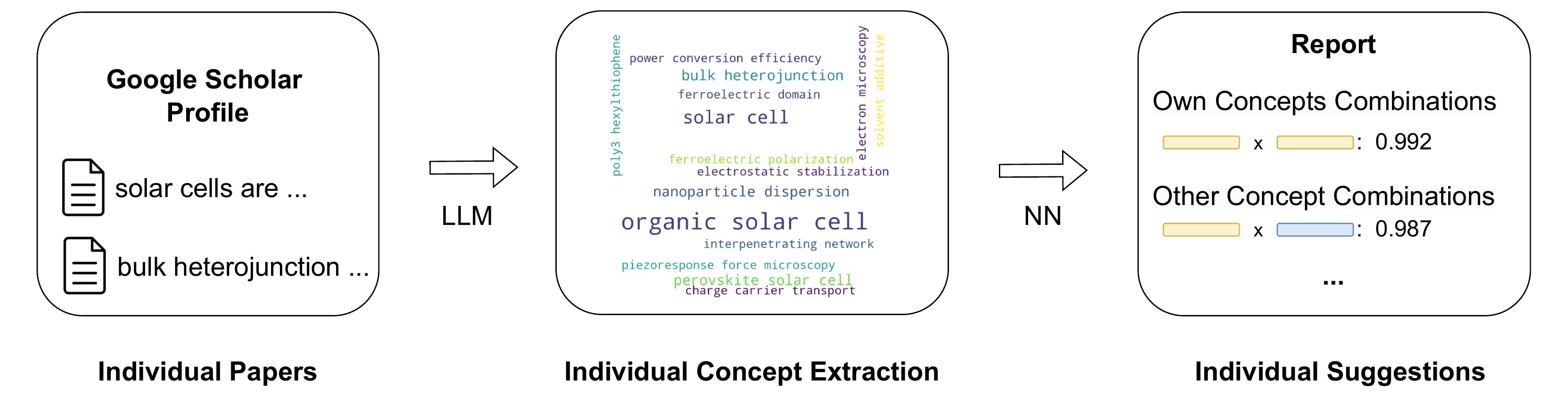}
    \caption{Overview of the report generation: (1) Selection of abstracts of recent publications, (2) Extraction of individual concepts using our fine-tuned LLM, and (3) Suggestion of combinations in a standardized report based on our best ML model for link prediction.}
    \label{fig:ed1-report-generation}
\end{figure}

\paragraph{Report generation.}
An overview of the report generation is shown in \ref{fig:ed1-report-generation}. First, a set of individualized concepts $C_{{\text{own}}}$ is generated as the intersection of (i) all concepts extracted from the abstracts of the recent publications of the respective researcher and (ii) all known concepts $C_{{\text{known}}}$ in the concept graph. Based on these two sets of concepts, we generated researcher-specific suggestions, i.e. combinations of concepts: The first two report sections, $S_{\text{own}\times\text{own}}$ and $S_{\text{own}\times\text{other}}$, contain the top 25 combinations of the own concepts with themselves and with all other concepts, respectively. We applied two heuristics (avoid generic concepts and avoid too similar as well as too unrelated combinations based on their semantic embeddings) to filter the suggestions in the second category, resulting in a section $S^{\text{filtered}}_{\text{own}\times\text{other}}$. 
To take the researcher's full profile into account, the next section $S_{\text{(many own)}\times\text{other}}$ contains the top 20 concepts with highly scored connections to many own concepts. For the final section of the report (`LLM curation'), LLMs were queried to select interesting combinations from the previous sets of combinations and to write a short paragraph with more information on how the concepts can be combined and why these specific combinations are promising new research directions.
A technical definition of each section is given in \autoref{si-sec:report}.

\paragraph{Classification of the suggestions.}
Based on the individual reports described above, a 30-minute interview was conducted with each of the researchers, in which the suggested combinations of concepts were classified as already known (A), nonsensical or not understandable (B), and novel, interesting, or inspiring (C). In section 4 ($S_{\text{(many own)}\times\text{other}}$), suggestions were generously already counted as overall interesting (category C) if one of the many concepts in $C_{\text{own}}$ was inspiring in conjunction with the proposed other concept. To account for cases in which the participants were unsure whether to label a suggestion as B or C, an additional category D was introduced in the analysis. 
For further analysis, the first class was divided into already published combinations (A1) which were likely missed during dataset generation (e.g., very recent publications or publications outside of the analyzed literature corpus) as well as obvious, trivial, or very general combinations (A2), which are not necessarily mentioned together in an abstract. 

The occurrence of the combinations classified as A1 (known), A2 (trivial), B (nonsense), C (interesting), D (uncertain) using this scheme is listed in \ref{tab:report-category-overview}. Out of 292 categorized suggestions, we find the following: Class A1: 71 suggestions, Class A2: 36 suggestions, Class B: 99 suggestions, Class C: 77 and Class D: 9, suggestions. Thus, 26\% of all suggested concept combinations were considered interesting by the interviewees. An excerpt of combinations per category can be found in \autoref{si-tab:ranked-combinations_a12} and \autoref{si-tab:ranked-combinations_bcd}. As mentioned above, the number of interview partners is not sufficient for a reliable statistical analysis, e.g. the total number of classifications per researcher ranged between 18 and 48, with a per-participant variance ranging from 5.04 to 51.36. An overview of the classified combinations per researcher and the per-participant variance can be found in (\autoref{si-fig:overview-concepts-per-researcher} and \autoref{si-tab:overview-concepts-per-researcher}. 

\begin{table}[ht]
    \centering\small
    \caption{Amount of suggestions categorized by researchers across all interviews, broken down by section of the report.}
    \begin{tabular}{C{.18\textwidth}C{.12\textwidth}C{.1\textwidth}C{.1\textwidth}}
        \toprule
        & &  \multicolumn{2}{c}{Occurance} \\ \cmidrule(lr){3-4}
        Section & Category & Number & Fraction\\\midrule
        \multirow{4}{*}{$S_{\text{own}\times\text{own}}$}  & A1 & 31 & 30.7\% \\
        & A2 & 21 & 20.8\% \\
        & B  & 25 & 24.8\% \\
        & C  & 21 & 20.8\% \\ 
        & D  & 3  & 3.0\% \\ \midrule
        \multirow{4}{*}{$S_{\text{own}\times\text{other}}$}
        & A1 & 17 & 22.1\% \\
        & A2 & 8  & 10.4\% \\
        & B  & 35 & 45.4\% \\
        & C  & 14 & 18.2\% \\
        & D  & 3  & 3.9\% \\ \midrule
        \multirow{4}{*}{$S_{\text{(many own)}\times\text{other}}$}
        & A1 & 6  & 23.1\% \\ 
        & A2 & 1  & \phantom{0}3.9\%  \\ 
        & B  & 3  & 11.5\% \\ 
        & C  & 16 & 61.5\% \\ \midrule
        \multirow{4}{*}{$S^{\text{filtered}}_{\text{own}\times\text{other}}$}
        & A1 & 17  & 19.3\% \\
        & A2 & 6   & \phantom{0}6.8\% \\
        & B  & 36  & 40.9\% \\ 
        & C  & 26  & 29.5\% \\
        & D  & 3   & 3.4\% \\ \bottomrule
    \end{tabular}
    \label{tab:report-category-overview}
\end{table}

To evaluate the usefulness of the `LLM Curation' approach, we analyze how many of the combinations suggested by an LLM are later labeled as interesting. In \autoref{si-table:llm-interesting-sugg-confusion-matrix}, we display the confusion matrix of the variables "Suggested by an LLM" and "Is Interesting", i.e. whether the suggestion was rated as C by a human expert.
Interestingly, we observe a rounded precision of 47\% regarding the LLM selecting interesting concepts, i.e. 24 of the 53 concepts suggested by the LLM as interesting were also labeled as interesting by the scientists. This is a substantial improvement in precision compared to 61 of 266 (23 \%) concept combinations analyzed in total. In this context the recall is not of primary interest, as the LLM was only allowed to select a limited number of combinations as described above. An analysis of the previous node distances of the suggested combinations across all reports showed that 5 out of 9 concept pairs at $d_{\text{prev}} = 3$, were rated as category `C'. This high ratio of interesting combinations underpins our previous assumption that the inclusion of semantic information in the prediction models increases its capability to foster out-of-the-box thinking. 

In retrospect, ground-breaking ideas sometimes seemed absurd at first. Interviewees repeatedly categorized a combination as B (`nonsensical') only to change their minds after some reconsideration, or when seeing that suggestion again in the `LLM Curation' section, together with an elaboration on how the combination could be realized. The exemplary paragraph often helped researchers to judge the usefulness of concept combinations. We speculate that the task of generating own hypotheses on how concepts could be connected is inherently more difficult than judging an existing proposal.

Furthermore, many combinations could not be classified, especially in the sections `Other Concept Combinations' and `Filtered Other Concept Combinations'. This is attributable to the vast amount of concepts in $C_\text{known}$, many of which the interviewees had never heard of. To allow researchers to navigate unknown suggestions, additional context, such as the original abstract, might prove helpful.

\subsection*{Human expert evaluation: Examples}
To illustrate the value of the suggested combinations of concepts in more detail, we discuss below five selected examples of category C suggestions by putting them into context and describing why these combinations are interesting. A more detailed discussion of all five concept combinations can be found in the Supporting Information.

\begin{figure}
    \centering\small
    \includegraphics[width=\textwidth]{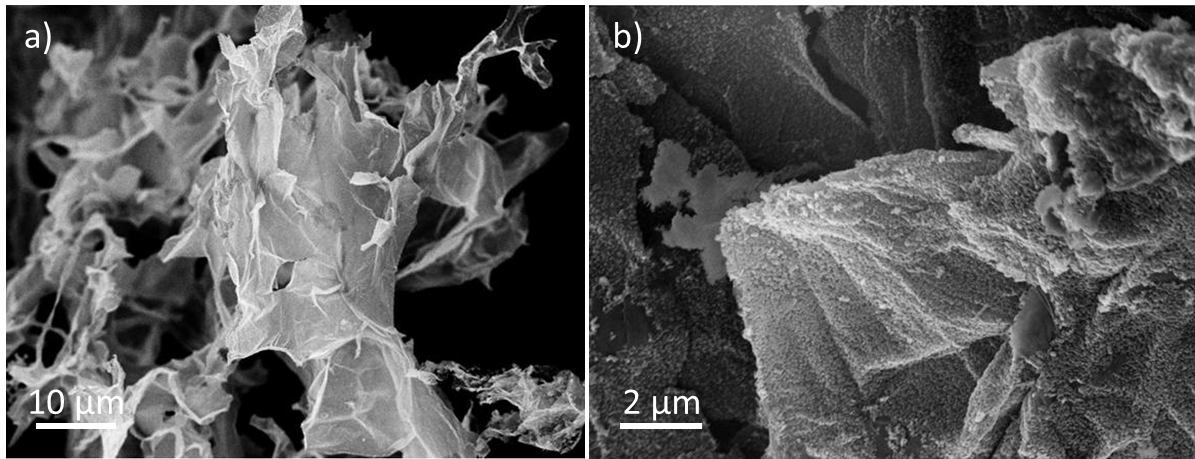}
    \caption{Exfoliated graphene oxide (resulting in multilayer graphene) covered with \SI{200}{nm} thick iron oxide layers. a) shows the flake morphology of exfoliated layers, b) shows the iron oxide layer covering the whole surface.}
    \label{fig:breitung}
\end{figure}

\subsubsection*{Suggestion: "Conventional ceramic" + "Graphene oxide"}
This suggestion associates "conventional ceramics" with "graphene oxide", two domains seldom combined. Conventional oxide ceramics provide chemical, thermal, and structural stability. Gra\-phene oxide offers a high-surface-area, electronically conductive carbon framework. Their union could yield composites marrying ceramic robustness with rapid charge and heat transport, relevant to batteries, catalysis, and thermal barriers. Existing studies mix pre-synthesised oxides with graphene derivatives, giving limited interfacial contact. We show preliminary data of a $\approx 200$ nm iron-oxide shell on multilayer graphene in \ref{fig:breitung} (unpublished). The in-situ process creates intimate oxide–graphene interfaces and a continuous conductive network. Electrochemical tests show high reversible capacity and enhanced redox kinetics in Li-ion conversion cells. These findings demonstrate the model's capacity to provide inspiration for overlooked but feasible synthesis strategies. Systematic exploration of AI-highlighted ceramic/graphene hybrids may accelerate multifunctional material discovery.

\subsubsection*{Suggestion: "Tensile strain" + "Molecular architecture"}
Thin-film organic and perovskite solar cells comprise multilayers with mismatched thermal-expansion coefficients. Temperature excursions during coating, annealing, and operation impose tensile strain at these organic–inorganic interfaces. Such strain drives delamination and point-defect formation, accelerating performance loss. While strain engineering is routine in inorganic semiconductors, it is rarely applied in soft-matter photovoltaics. Molecular architecture provides a complementary lever: greater torsional flexibility lowers film modulus and dissipates stress. Thus, the AI-proposed link “tensile strain + molecular architecture” highlights an under-exploited stability pathway. In 2024, it was shown by Brabec \& Friederich that TPA-containing hole-transport layers benefit efficiency by accommodating strain\cite{wu2024inverse}. These studies corroborate strain-aware molecular design as a broadly applicable interface strategy. Systematic exploration could extend device lifetimes without compromising performance.

\subsubsection*{Suggestion: "Multiphase structure" + "Selective laser melting"}
Microstructure denotes all internal structural features—from lattice arrangement to point, line, planar, and volumetric defects—across relevant length scales. These features dictate mechanical and functional response and thus underpin rational materials selection. A central attribute is the spatial distribution of phases, each possessing uniform crystal structure and composition. Technical alloys and ceramics are typically multiphase, generated through controlled thermo-mechanical processing. Phase topology manipulation enables simultaneous optimization of strength, toughness, corrosion resistance, and functional properties. Selective Laser Melting (SLM) fabricates components by layer-wise laser melting of metal powders directly from digital models. The extreme heating–cooling rates inherent to SLM impose strong non-equilibrium solidification conditions. Resulting parts frequently exhibit metastable, compositionally heterogeneous multiphase microstructures. These structures can elevate hardness and corrosion resistance, yet they may also induce residual stresses. Therefore, elucidating phase-formation pathways during SLM represents a critical avenue for advanced materials design.

\subsubsection*{Suggestion: "Stress induced phase transformation" + "Hexagonal boron nitride"}
Stress-induced phase-transformation toughening, exemplified by the tetragonal to monoclinic switch in zirconia, suppresses crack advance through local volume expansion. An alternative route uses elastic anisotropy; in pearlitic wires, load-parallel micro-cracks raise both strength and toughness. Applying these principles to boron nitride asks whether hexagonal BN (h-BN) can function as a transformation- or anisotropy-assisted toughener. Cubic BN (c-BN) is a dense, superhard phase, and pressure-driven c-BN to h-BN transitions could release crack-tip stresses. h-BN displays strong in-plane vs. out-of-plane stiffness contrast, enabling guided micro-crack arrays akin to the pearlite mechanism. A coupled h-BN anisotropy and c-BN/h-BN transformation would thus offer simultaneous crack deflection and compressive shielding. Recent c-BN/h-BN composites show higher hardness and fracture energy, indicating technological promise. However, the role of a reversible transformation in these gains remains experimentally unverified. Targeted high-pressure mechanical tests with in-situ diffraction are required to resolve transformation kinetics and toughening contributions. Establishing these links could generalize anisotropy-assisted transformation toughening to lightweight nitride coatings.

\subsubsection*{Suggestion: "In-plane polarization" + "Organic solar cell"}
Ferroelectric in-plane polarization, recently demonstrated in MAPbI$_3$ perovskites, spatially separates photocarriers and channels them toward electrodes. The ML suggestion indicates that similar lateral dipole fields could be engineered in organic absorbers. Asymmetric polar moieties, oriented during self-assembly or within covalent organic frameworks, may supply the required non-centrosymmetry. The resulting internal field should enhance carrier separation and transport, while a higher dielectric constant lowers exciton binding and monomolecular recombination. Ferroelectricity is so far verified only for halide perovskites \cite{rossi_ferroelectric_2018,rohm_ferroelectric_2017,schulz_ferroelectricity_2022}, and is absent in silicon or conventional organic cells \cite{rohm_ferroelectric_2019,breternitz_role_2020}. Piezoelectric polymers such as PVDF already exploit oriented dipoles in sensing devices, suggesting viable processing routes. Prior attempts to raise organic permittivity show limited success, and deliberate in-plane polarization remains unexplored \cite{roehm2020}. Hence, the predicted concept defines a tractable, novel direction for photovoltaic materials research.

\section*{Discussion}
In the first part of this study, we showed that the power of large language models, especially LLaMa-2-13B, can be harnessed to extract scientific concepts -- vaguely defined as key phrases -- from scientific text. We established a methodology for fine-tuning open-source LLMs based on a small manually labeled abstracts, which guides the LLM to extract only relevant concepts. The initial training data can be iteratively extended by humanly corrected LLM annotations to further improve the extraction process, but no human verification is required to check the final 221,000 labeled data points. Follow-up studies may investigate whether prioritizing quality over quantity \cite{zhou_lima_2023} in the annotated training examples, i.e. by using fewer but carefully selected data points, could yield more accurate and representative extracted concepts, and also whether the inclusion of synthetic data can help to accelerate the annotation process and further enhance model performance. 

To achieve the second goal of this work, we created a concept graph, derived from the previously extracted materials science concepts and the dates of the corresponding publications. This graph was then successfully used to predict emerging links between previously unconnected concepts, underscoring that a simple graph representation suffices for this task. Finally, we demonstrated that the integration of semantic knowledge in the form of concept embeddings boosts the predictive performance of our model. Combining the GNN approach with semantic features is possible and will be explored in future work. The usefulness of our model in a real-world scenario was investigated through qualitative interviews with domain experts, who rated 77 or 26\% out of 292 generated recommendations as interesting. While this rate may sound modest, each 30-minute session still yielded several promising ideas, making the outcome practical for guiding research.

In summary, we demonstrated that machine learning tools can be used to automatically process the vast amount of scientific literature and to predict future research directions that have not been explored before to foster innovation and advancements. While this work focused on the material sciences as a use case, the developed approach can easily be extended to other research areas. By suggesting potential new research directions, we hope to drive innovation and collaboration in the field.

% end of main text

%TC:ignore
\section*{Methods}
\label{sec:methods}
\begin{figure}
    \centering\small
    \includegraphics[width=\dimexpr1.0\textwidth\relax]{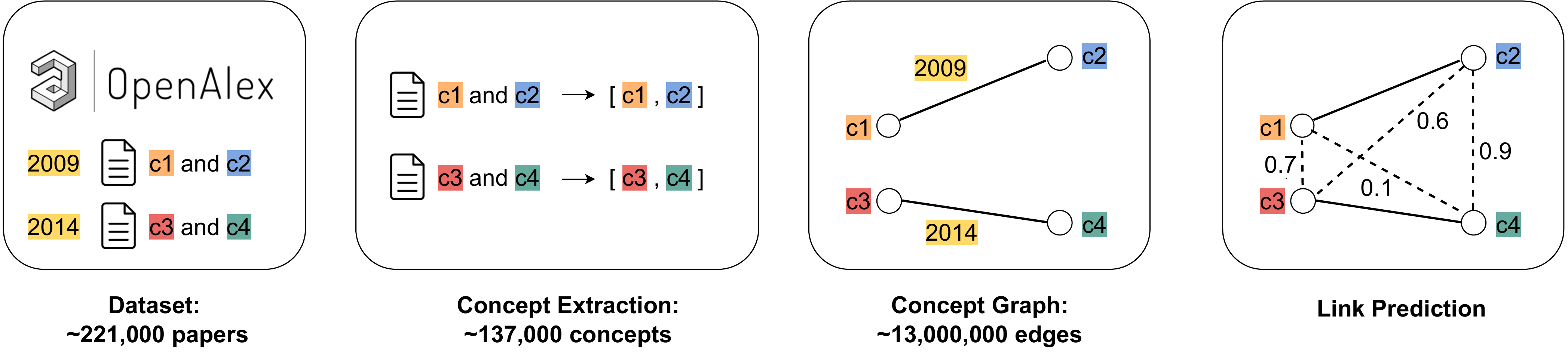}
    \caption{Overview of the link prediction workflow: (a) gathering of materials science abstracts with OpenAlex \cite{priem_openalex_2022}, (b) extraction of concepts using LLMs, (c) creation of the semantics-aware-concept graph, and (d) prediction of new research directions.}
    \label{fig:ed-fig3}
\end{figure}

The key steps of our approach are depicted in \ref{fig:ed-fig3}. After gathering the abstracts of a large number of research publications in the domain of materials sciences, we extracted the main concepts, i.e. short key phrases consisting of few words from these abstracts, and used them as nodes in a concept graph that mirrors the (time-dependent) connectivity of the materials science concepts in literature. The final step of our workflow consists of performing link prediction on this graph based on both network properties, e.g. connectivity information, and semantic knowledge about the concepts captured in aggregated word embeddings.

\subsection*{Dataset}
We prepared a dataset of published papers related to materials science. Data was obtained from OpenAlex by querying all publications listed at materials science-related journals, conferences, and other venues \cite{priem_openalex_2022}. The retrieved papers were filtered based on language, length, and whether they have an abstract. For each publication, the title and abstract were cleaned and concatenated. Chemical formulae were extracted, stored separately, and later merged with the extracted concepts. The resulting dataset comprises approximately 221,000 articles published between 1955 and 2022, with relevant attributes being "title", "abstract", and "publication date". A more detailed description of the dataset generation is given in \ref{si-sec:dataset}.

\subsection*{Concept extraction}
\label{sec:concept-extraction}

\begin{figure}
    \centering\small
    \includegraphics[width=0.7\textwidth]{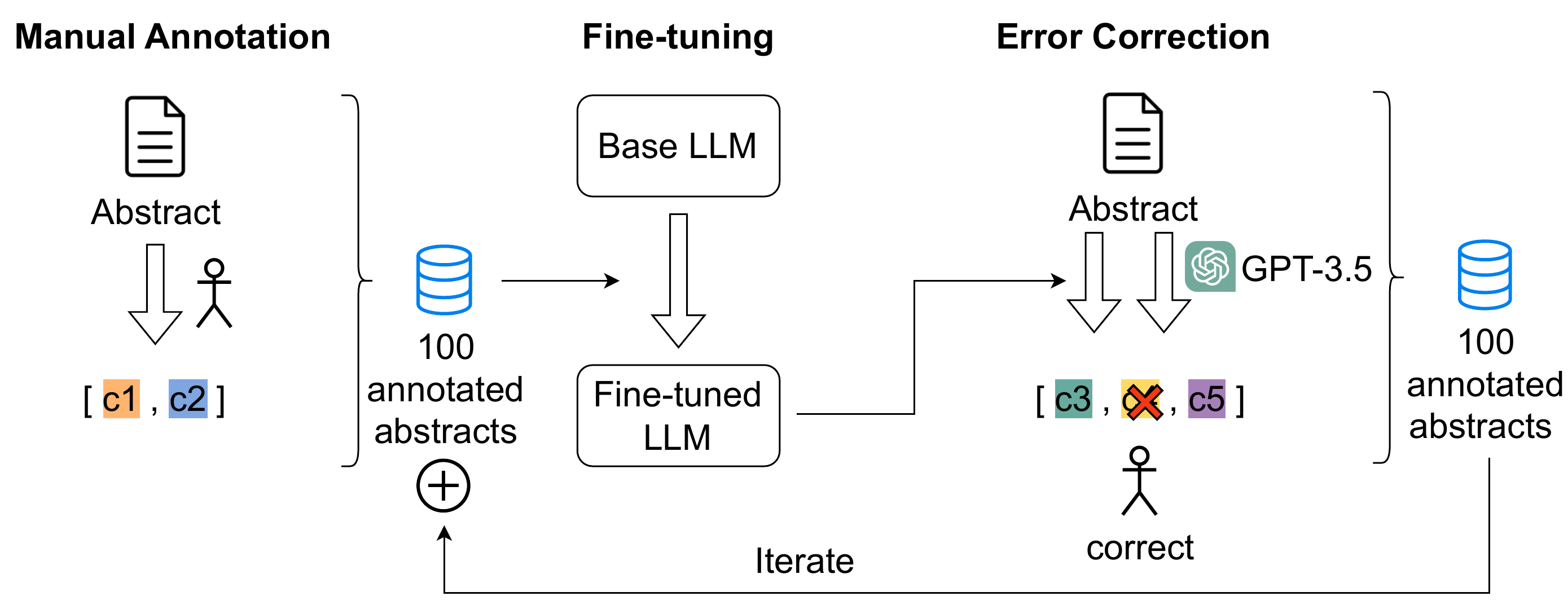}
    \caption{Generation of labeled data: (1) manual labeling (concept extraction) of 100 abstracts, (2) fine-tune a LLM-base model on the annotated data, (3) automatic concept extraction from 100 further abstracts with human correction, and (4) repetition of fine-tuning of the base LM with the new extended labeled dataset.
    }
    \label{fig:generate-training-data}
\end{figure}

Previous work used RAKE for concept extraction in conjunction with manual filtering to remove errors, i.e. phrases that do not represent semantic information, introduced by the imperfect statistical analysis performed in RAKE \cite{krenn_predicting_2020, krenn_predicting_2022, rose_automatic_2010}. Instead, we opted for extracting concepts using fine-tuned LLMs (see \autoref{fig:generate-training-data}). To create a dataset for fine-tuning, 100 randomly chosen abstracts were first manually annotated by extracting and partially adjusting or even paraphrasing relevant and meaningful concepts as a preliminary step. We note that manual annotation is particularly sensitive to the labeler, as there is no unique way of extracting and defining concepts.
Subsequently, we fine-tuned the LLaMa-2-13B base model \cite{touvron_llama_2023,touvron_llama_2023-1} on our manually annotated abstracts for 4 epochs, using a learning rate of $5 \cdot 10^{-4}$ and a batch size of 1  (see \ref{si-sec:hyperparameters}). Llama-2 models were SOTA when the work was performed in 2023, but will be replaced with newer models in future iterations of this work). The size of the model is a trade-off between accuracy and cost, as it is the largest model that can process 20 abstracts at once on an A100 GPU with 80GB of VRAM. To accelerate training and especially inference, we incorporated 8-bit quantization \cite{dettmers_llmint8_2022} and low-rank adaptation (LoRa) techniques \cite{hu_lora_2021, dettmers2023qlora} using Hugging Face's PEFT module \cite{peft}. 

Similar to Dunn et al.'s assisted annotation process \cite{dunn_structured_2022}, the fine-tuned model's outputs were compared in the third step to the concepts extracted by GPT-3.5 \cite{brown_language_2020} to efficiently identify and correct common mistakes of the fine-tuned model. In this way, we labeled 100 additional abstracts, and the base model was again fine-tuned on the larger dataset of 200 abstracts. The process of iteratively adding more automatically extracted and manually corrected concepts to fine-tune the model with more datapoints could, in principle, be repeated more often, but 200 labeled abstracts were enough for our use case. Finally, the resulting model was employed to extract concepts from all approximately 221,000 abstracts in our dataset, requiring approximately 160~GPU-hours. Future updates of our concept graph are substantially less demanding, as only incremental (delta) extraction is required. Future developments in LLM research might enable a complete re-evaluation with higher quality and reliability.  After extraction, we conducted minor post-processing of the extracted concepts by removing the remaining plural forms. We note that some bias was introduced by the selection of 200 abstracts with 3102 distinct concepts, as they were all from materials science,  and observed that, e.g., core-biology concepts were not extracted by our method.

\subsection*{Concept graph}
\label{sec:concept-graph}

The concepts extracted from the materials science literature in our dataset are represented in a multi-graph \( G = (V, E) \) where \(V\) and \(E\) are the respective sets of nodes and edges. In this concept graph, each node \( v \in V \) represents a distinct concept, and each edge \( e \in E \) represents the co-occurrence of two concepts in a single abstract. Each edge is labeled with a timestamp \(t\), indicating the publication date of the abstract containing both of the two concepts where $G_t$ denotes a subgraph of $G$ defined by the edge list that only includes all edges with time stamps $\leq t$.
Therefore, every abstract generates a fully connected clique of its concepts in the concept graph. Multiple edges can exist between each pair of nodes if the concepts co-occurred in more than one abstract.

\begin{figure}
    \centering\small
    \includegraphics[width=\textwidth]{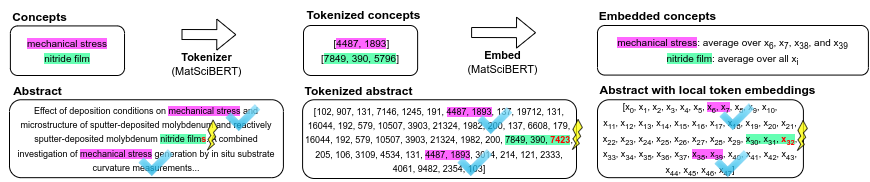}
    \caption{Example for calculating concept embeddings from an abstract. Embeddings of verbatim concepts ('mechanical stress') are calculated by averaging over all local MatSciBert embeddings of the corresponding tokens (4487 and 1893). Embeddings of non-verbatim concepts ('nitride film' is only present in the abstract in its unnormalized form 'nitride films') are calculated as the average over all token embeddings.}
    \label{fig:extract-embeddings}
\end{figure}

To enrich the nodes of our concept graph with semantic information, we calculate concept embeddings and use them as node features.
The procedure of calculating concept embeddings using MatSciBERT \cite{gupta_matscibert_2021} is summarized in \autoref{fig:extract-embeddings}: First, both the entire abstract and the previously extracted concepts are tokenized, and the embeddings are then calculated for the tokenized abstract. The next step consists of locating all instances of a concept in the abstract and averaging the embeddings of the tokens corresponding to the concept. For example, the concept ‘mechanical stress’ is tokenized as [4487, 1893], and the concept embedding is calculated as the average of the corresponding representations in the embedded abstract, i.e. those corresponding to the positions of the sequence [4487, 1893] in the tokenized abstract. To derive a singular representation for each concept per abstract, we average the embeddings of all its occurrences. In cases where a concept does not appear verbatim in an abstract — e.g. due to the normalization processes during the initial concept extraction — we take the mean embedding of all tokens (while excluding the start and end token in the abstract as its representation). As the final step, the average embedding for identical concepts across different abstracts is calculated to obtain a single embedding for each concept, and thus for each node. To prevent information leakage, embeddings used for training and testing are computed only from text available up to the corresponding cutoff year.

\subsection*{Link prediction}

The previous method used by Krenn \emph{et al.} to predict new links in their concept graph exclusively relied on abstract local graph properties, either through a purely graph-theoretical approach using hand-crafted features in conjunction with machine learning or through employing end-to-end machine learning methods \cite{krenn_predicting_2020,krenn_predicting_2022}. Here, we investigate whether integrating semantic knowledge about the concepts can improve link prediction. In particular, we use concept embeddings, i.e., high-dimensional vectors that capture semantic information, to make the semantic information integrable into the link prediction task.

\begin{figure}
    \centering\small
    \includegraphics[width=\textwidth]{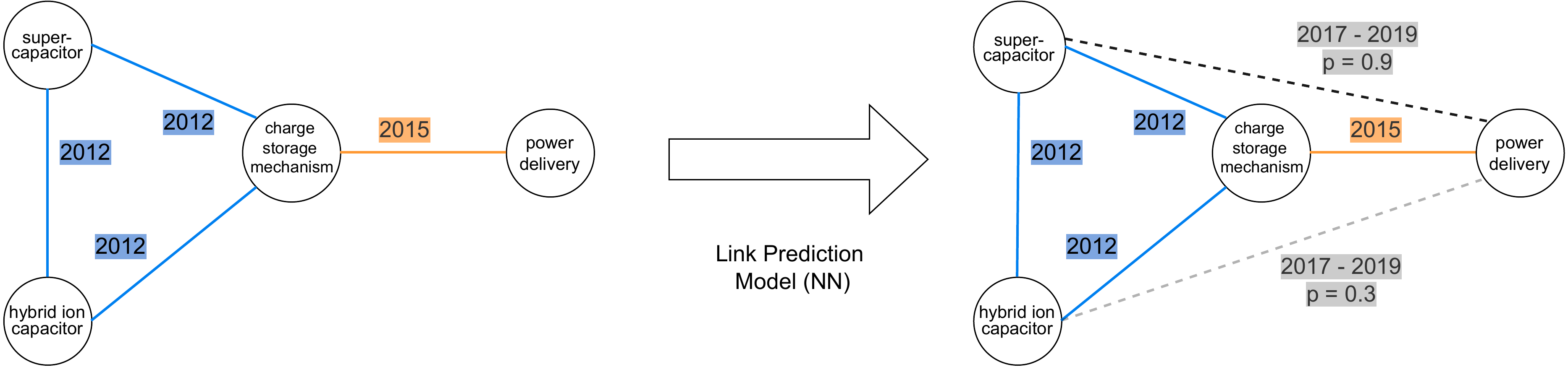}
    \caption{Example for a concept graph derived from two articles published 2012 (blue) in and 2015 (orange) with one overlapping concept. Possible new edges are marked as dotted gray lines together with their predicted probability of formation in the years 2017 to 2019.
    }
    \label{fig:concept-graph}
\end{figure}

Given the concept graph $G$, we treat link prediction as a binary classification task. The objective of the ML model is thus to predict whether a new edge is formed between an arbitrary pair of previously unconnected vertices $(u, v)$ in the time range $T = [T_{\mathrm{start}}, T_{\mathrm{end}}]$. We chose $T_{\mathrm{start, train}} = 2017$ and  $T_{\mathrm{end, train}} = 2019$ for training, i.e. our model has access to the entire data up to and including 2016 while its predictions are made for the years 2017, 2018, and 2019. We illustrate link prediction on a rudimentary concept graph in \ref{fig:concept-graph}.

The prediction task has an inherent strong label imbalance since the likelihood of a randomly selected vertex pair $(u, v)$ forming a link throughout three years is extremely low. While there are for example 18.7 billion possible new edges that could form between 2017 and 2019, only 1.3 million new edges ($0.007\%$) can actually be observed during this period. To address this imbalance, we oversampled positive labels by using a fixed percentage (30\%) of positive examples per batch in the training process. This oversampling \emph{during training} shifts the trade-off between precision and recall in imbalanced tasks toward higher recall at the cost of losing precision, thus favoring the generation of larger sets of suggestions that may contain inspiring concept combinations over smaller sets in which some valuable ideas might not be included.

A modified version of Krenn \emph{et al.}'s densely connected neural network \cite{krenn_predicting_2022}, which purely relies on graph properties at different points in time, was used as the \emph{Baseline} model.
Specifically, the degree of a node $u$ ($\sum_{i=1}^{n} A_{u,i}$) and the sum of all 2-length paths from $u$ ($\sum_{i=1}^{n} A_{u,i}^{2}$) were calculated for different years in the range $t = [T_{\text{start,train}}-5,T_{\text{start,train}}-1]$, where $A_t$ denotes the binary adjacency matrix of $G_t$.
These features were then concatenated for a given pair of nodes $(u,v)$ to result in a 20-dimensional baseline feature vector. In the second \emph{Concept embeddings (MatSciBERT)} model, the concatenated concept embeddings of $u$ and $v$ were used instead as the (1,536-dimensional) feature vector for the neural network classifier, to test their information content and relevance for the link prediction task. We repeat the embeddings generation process with BERT, yielding a modified \emph{Concept embeddings (BERT)} model. To explore another way of utilizing semantic information, we fine-tune the MatSciBERT model directly to predict the likelihood of two concepts becoming connected in our concept graph, thus, yielding the \emph{Pure Text Baseline} model.

The \emph{Baseline} model was then combined with the \emph{Concept embeddings} model in two hybrid
models, for the first of which (\emph{Combination of features}), a concatenation of the feature vectors of the two previous models was used as the input. The hyperparameters of all neural networks were optimized using a comprehensive grid search varying the number of neurons per layer, the percentage of positive samples in each batch, the learning rate, and the dropout probability (see \autoref{si-tab:hyperparameters} for a list of optimized hyperparameters).

The second hybrid model (\emph{Mixture of Baseline and Embeddings}) uses a weighted output of the optimized \emph{Baseline} and \emph{Concept embeddings (MatSciBERT)} models where an optimal weighting of 3:2 was determined using hyperparameter optimization. We note that there are many other possible hybrid models beyond concatenating the two input vectors and calculating the weighted average of the output probabilities. The two parts of the input could be passed through a first set of layers separately before the
two outputs are concatenated and passed through a second set of layers. The optimization of the
architecture of the NN was however outside the scope of this study, as our goal mainly consisted
in showing that the inclusion of the concept embeddings improves link prediction.

To explicitly capture local neighborhood structures through message passing, we implemented a graph neuralnetwork (\emph{GNN}) model. Given the large-scale and hub-heavy nature of the network, we employed neighbor sampling to enable efficient training, initializing the node representations with the topological vectors from the \emph{Baseline} model. This architecture utilizes a 2-layer GraphSAGE encoder \cite{hamilton_inductive_2017} with neighbor sampling to compute node embeddings based on the graph topology at $T_{\text{start,train}}$. An MLP decoder was employed to classify the concatenated node embeddings.

Finally, we constructed a \emph{Mixture of GNN and Embeddings} model as a third hybrid model, analogous to the \emph{Mixture of Baseline and Embeddings} approach described above. This ensemble calculates a weighted average (1:1) of the output probabilities from the \emph{GNN} and the \emph{Concept embeddings (MatSciBERT)} models.

To avoid overfitting, we monitor the area under the curve (AUC) -- a summary metric we derive from the Receiver Operator Characteristic (ROC) -- on a potentially out-of-distribution validation set with $T_{\mathrm{start,validation}} = 2020$ and $T_{\mathrm{end,validation}} = 2022$.

\subsection*{Human domain experts}
The human domain experts who participated in the interviews are affiliated with different institutes at different institutions and were recruited to cover a wide range of topics within materials science. 13 researchers were invited to participate, 10 of which agreed to participate in the study and to conduct interviews. All participants are professors or independent group leaders. No interviews were excluded from the study.

\section*{Code availability}
The code used to produce the results of this study is publicly available and permanently archived on Zenodo \cite{materials_concepts}. The source code is licensed under the GPL 3.0+ License and the active development repository can be found at \url{https://github.com/aimat-lab/materials_concepts}.

\section*{Acknowledgements}
Creating a chemical element parser from scratch is hard. We thank Chris Konop for an outline of how a chemical formula can be recursively defined. This made our implementation way more straightforward.
We acknowledge support by the Federal Ministry of Education and Research (BMBF) under Grant No. 01DM21001B (German-Canadian Materials Acceleration Center) (P.F.).
We acknowledge funding by the German Research Foundation (DFG) under Germany's Excellence Strategy via the Excellence Cluster “3D Matter Made to Order” (3DMM2O, EXC-2082/1–390761711) (P.F.).
Funding by the German Research Foundation (DFG) through the collaborative research center CRC 1249 N-Heteropolycycles as Functional Materials (SFB 1249, Project C13) and through DFG Projekt 436506789 is gratefully acknowledged (P.F.). 
This work was partly carried out with the support of the Karlsruhe Nano Micro Facility (KNMFi, www.knmf.kit.edu), a Helmholtz Research Infrastructure at Karlsruhe Institute of Technology (KIT, www.kit.edu) (B.B.).
We acknowledge funding from the Helmholtz Metadata Collaboration (HMC) through the project AIMWORKS (P.F.).
This work was performed on the HoreKa supercomputer funded by the Ministry of Science, Research and the Arts Baden-Württemberg and by the Federal Ministry of Education and Research. The authors acknowledge support by the state of Baden-Württemberg through bwHPC (P.F.).

\pagebreak

\setcounter{section}{0}

\renewcommand{\thesection}{S\arabic{section}}

\begin{center}
    \textbf{\huge Supplementary Information}
\end{center}
\vspace{1em}

\section{Concept extraction}
\subsubsection*{Evaluating LLM-based Concept Extraction}
\label{si-sec:llm-concept-evaluation}

\begin{sifigure}[b!]
    \centering\small
    \includegraphics[width=\textwidth]{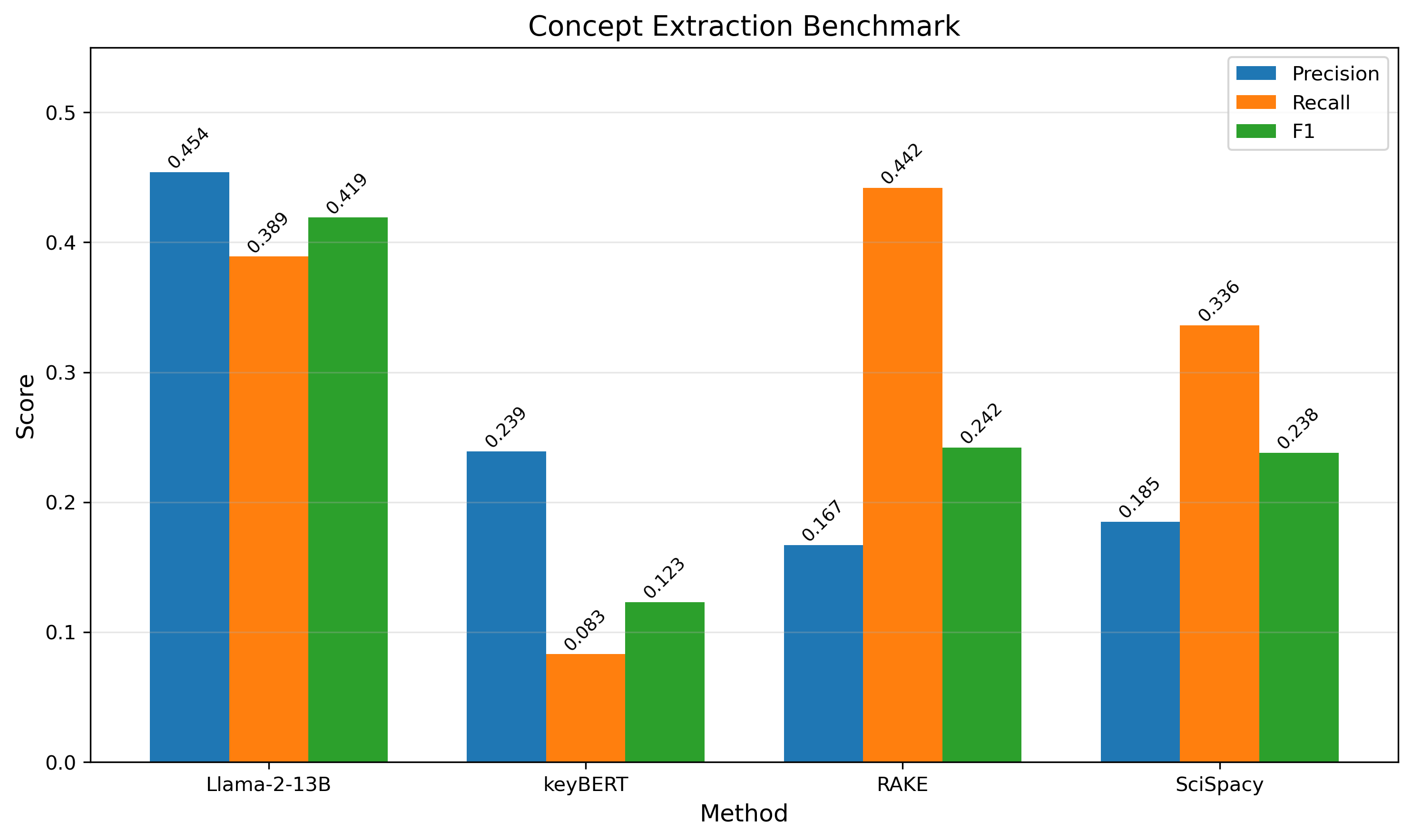}
    \caption{Precision, Recall and F1-Score of our LLM-based approach, keyBERT, RAKE and SciSpacy; evaluated against a held-out test set of 5 annotated abstracts, including 138 ground-truth concepts in total.}
    \label{si-fig:concept-extraction-benchmark}
\end{sifigure}

To construct a held-out test set, three annotators (n=3) manually annotated five abstracts that were not part of the training data. The resulting inter-annotator agreement, measured as the mean pairwise agreement, was 0.345 (pairwise scores: 0.215, 0.326, 0.493; all values rounded to three decimal places).

Importantly, none of the human-annotated concepts were incorrect. Disagreement primarily stemmed from varying emphasis on annotation breadth versus depth. As all extracted concepts were valid, we defined the union of all annotations as the ground truth for evaluating our LLM-based concept extraction approach against established methods, namely KeyBERT, RAKE, and SciSpacy.

RAKE achieved slightly higher recall than the LLM-based approach, primarily due to its extensive extraction of candidate terms. However, this also resulted in substantially lower precision, as for every relevant concept, approximately five irrelevant ones were produced (precision $\approx16.7\%$). When balancing precision and recall, the LLM-based method outperformed all other approaches, achieving a substantially higher precision at similar recall compared to RAKE.

Notably, concepts produced by the LLM that were not counted as true positives tended to be semantically relevant but either overly broad or overly specific (e.g., `aerospace part thickness measurement'), whereas competing methods frequently generated entirely irrelevant outputs such as `6148 10 3 cycles', `increase', `interdisciplinary field involving principles', or `317'. This highlights the practical advantage of the LLM-based approach: it substantially reduces the need for manual post-processing, as the extracted concepts are largely free of irrelevant entries.

To enable researchers to benchmark their own models, all labeled data can be found on our GitHub repository.
\newpage
\subsubsection*{Examples of extracted concepts}
\newcommand{\conceptlst}[5]{
    \begin{sifigure}[ht!]
    \small
        \fbox{\parbox{\linewidth}{
        \setlength{\fboxsep}{0pt}
        \textbf{Abstract}\\
        #1\\
        
        \begin{tabular}{p{0.5\linewidth}p{0.5\linewidth}}%\toprule
            \textbf{LLM Concepts} & \textbf{Rake Concepts} \\ \midrule
            \makecell[l]{#2} & \makecell[l]{#3} \\
        \end{tabular}
        \setlength{\fboxsep}{3pt}
        }}
        \caption{#4}
        \ifthenelse{\equal{#5}{1}}{\label{table:concept-extraction-1}}{}%
        \ifthenelse{\equal{#5}{6}}{\label{table:concept-extraction-6}}{}%
    \end{sifigure}
}

\conceptlst{ % ABSTRACT
\colorbox{yellow}{Removal of Metal Impurities} from the Spent Hydroprocessing Catalyst and Acid Leaching Kinetics of Aluminum. The impurity metal of Al in spent hydroprocessing catalyst MoNi/Al2O3 was removed by HCl, and the metals of Mo and Ni were preliminarily enriched. The results show that the leaching efficiency of Al was 88.62\%, and the leaching efficiency of Mo and Ni were 16.32\% and 28.74\%, respectively. The results were achieved under optimal leaching conditions: the particle size was 150 m, the concentration of HCl was 4 mol/L, the leaching temperature was, and the leaching time was 120 min. The kinetics of the leaching behavior of Al showed that the \colorbox{yellow}{acid leaching reaction of Al }was in accorded with the equation 1-2/3/X-1-X 2/3/ =K 2 t, R 2 =0.97734, which was controlled by internal diffusion. X-ray diffraction analysis of the leaching residue revealed the existence of residual metals MoMoO3 , MoO2ClO42 and NiNiS. The separation of Al from Mo and Ni has been preliminarily realized, which is conducive to further efficient recovery of Mo and Ni.
}{ % CONCEPTS
acid leaching kinetic \\
\colorbox{yellow}{aluminium acid leaching} \\
aluminium impurity \\
aluminium removal \\
hydroprocessing catalyst \\
hydroprocessing catalyst impurity \\
metal impurity \\
\colorbox{yellow}{metal impurity removal} \\
spent hydroprocessing catalyst \\
x ray diffraction analysis
}{ % RAKE CONCEPTS
acid leaching kinetics \\
acid leaching reaction \\
al showed \\
efficient recovery \\
equation xx \\
impurity metal \\
internal diffusion xray diffraction analysis \\
leaching behavior \\
leaching efficiency \\
leaching efficiency \\
leaching residue revealed \\
leaching temperature \\
leaching time \\
metal impurities \\
optimal leaching conditions \\
particle size \\
preliminarily enriched \\
preliminarily realized \\
residual metals momoo mooclo \\
results show \\
spent hydroprocessing catalyst \\
spent hydroprocessing catalyst monialo \\
}{ % CAPTION
Concept Extraction for W3027426609
}{1} % LABEL

\conceptlst{ % ABSTRACT
Strengthening mechanisms in short \colorbox{yellow}{carbon fiber reinforced} Nb/Nb5Si3 composites with interfacial reaction. The understanding of strengthening mechanisms in composites is crucial to improve its mechanical properties. Strengthening mechanisms in Nb/Nb5Si3 composites reinforced by different contents of short carbon fiber C sf were investigated. Interface and tensile properties of the composites were studied. Interfacial nano-sized Nb3 phase forms by in situ interfacial reaction, the Nb3 phase embedded intimately in the C sf and matrix forms an anchor effect. The content of Nb3 phases in the composites increases because of the increase of C sf content. Ultimate tensile strength and yield strength of the composite are improved because of the increase of C sf content. When the content of C sf is 1.0 wt\%, the elongation reaches the maximum, but the elongation decreases slightly when the content of C sf increases from 1.0 wt \% to 2.0 wt \%. The strengthening mechanisms in this composite were investigated according to grain refinement and load transfer. Load transfer is main strengthening mechanism and accounts for more than 95\% of the improvement in yield strength. This indicates that the control of interfacial bonding state is crucial to improve the mechanical properties of the composites. This study can offer some valuable information to disclosing the strengthening behaviors in the C sf reinforced metal matrix composites.
}{ % CONCEPTS
anchor effect \\
\colorbox{yellow}{carbon fiber reinforcement} \\
composite strengthening mechanism \\
elongation \\
grain refinement \\
in situ interfacial reaction \\
interface property \\
interfacial reaction \\
load transfer \\
mechanical property \\
nano sized phase \\
short carbon fiber \\
strengthening mechanism \\
tensile property \\
ultimate tensile strength \\
yield strength
}{ % RAKE CONCEPTS
anchor effect \\
c sf \\
c sf content \\
c sf content ultimate tensile strength \\
c sf increases \\
c sf reinforced metal matrix composites \\
composites increases \\
different contents \\
elongation decreases slightly \\
elongation reaches \\
grain refinement \\
interfacial bonding state \\
interfacial reaction \\
investigated according \\
investigated interface \\
load transfer load transfer \\
main strengthening mechanism \\
matrix forms \\
mechanical properties \\
mechanical properties strengthening mechanisms  \\
nb phase embedded intimately \\
nb phases \\
nbnbsi composites reinforced \\
short carbon fiber c sf \\
short carbon fiber reinforced nbnbsi composites \\
situ interfacial reaction \\
strengthening behaviors \\
strengthening mechanisms \\
studied interfacial nanosized nb phase forms \\
tensile properties \\
valuable information \\
yield strength \\
}{ % CAPTION
Concept Extraction for W3192967258 (normalization of concepts through LLM extraction).
}{2} % LABEL

\conceptlst{ % ABSTRACT
Gas-Channel Design of Gas-Assisted Molded Parts. The structure design of parts in \colorbox{yellow}{gas assisted injection moldingGAIM} was researched,while the parts with rectangular and semi-circular gas channels were studied by the methods of numerical simulation and physical simulation.At last,the degree of gas fingering was reduced numerical simulating experiment,CATIA was used to design simulated parts,and Moldflow was used to simulate the flow of plastic in the mold of GAIM.In physical simulating experiment,the dimension of gas-assisted injected plate was 160 mm140 mm.Gas channels were decussated in the plate.Gas channels with different dimensions were designed to obtain the effects of gas channels on gas fingering.The results show that in the design of semi-circular gas channel there is optimized proportion range of radius to plate thickness and the gas fingering of parts with square gas channels is serious.
}{ % CONCEPTS
catia \\
flow of plastic \\
\colorbox{yellow}{gas assisted injection molding} \\
gas assisted molded parts \\
gas channel design \\
gas channel dimension \\
gas fingering \\
gas injected plate \\
moldflow \\
numerical simulation \\
physical simulation
}{ % RAKE CONCEPTS
design simulated parts \\
different dimensions \\
gas assisted injection moldinggaim \\
gas channels \\
gas fingering \\
gasassisted injected plate \\
gasassisted molded parts \\
gaschannel design \\
mm mm gas channels \\
numerical simulating experiment catia \\
numerical simulation \\
optimized proportion range \\
physical simulating experiment \\
physical simulation \\
plate gas channels \\
plate thickness \\
results show \\
semicircular gas channel \\
semicircular gas channels \\
square gas channels \\
structure design
}{ % CAPTION
Concept Extraction for W2375662644.
}{3} % LABEL

\conceptlst{ % ABSTRACT
Microstructure and Wear Resistance of \colorbox{yellow}{Al2O3Coatings} on Functional Structure. To enhance the wear properties of function structure, Al2O3 -13\%TiO 2 AT13 coatings were plasma sprayed on 45 steel functional structure using micro and nano powders. The microstructures and phase compositions of the coatings were investigated by scanning electron microscopy and X-ray diffraction, respectively. Results show that the nano powder coating consists of fully-melted region and partially-melted region. The fully-melted regions show a lamellar structure, while the partially-melted regions retain the powders structure. The phases of coatings are -A1 2 O3 and TiO2 .The wear test was carried out on a ML-10 friction and wear tester under dry sliding condition. It is found that the wear resistance of the micro powder coating is higher than that of nano powder coating. This is mainly ascribe to the breakage of the nano powder coating resulted from low agglomerated binding force.
}{ % CONCEPTS
agglomerated binding force \\
\colorbox{yellow}{al2o3 coating} \\
coating microstructure \\
dry sliding condition \\
friction and wear tester \\
functional structure \\
lamellar structure \\
partially melted region \\
phase composition \\
plasma spraying \\
scanning electron microscopy \\
wear property \\
wear resistance \\
wear tester \\
x ray diffraction
}{ % RAKE CONCEPTS
dry sliding condition \\
fullymelted region \\
fullymelted regions show \\
function structure alo tio \\
functional structure \\
lamellar structure \\
low agglomerated binding force \\
mainly ascribe \\
micro powder coating \\
ml friction \\
nano powder coating \\
nano powder coating consists \\
nano powder coating resulted \\
nano powders \\
partiallymelted region \\
partiallymelted regions retain \\
phase compositions \\
plasma sprayed \\
powders structure \\
scanning electron microscopy \\
steel functional structure using micro \\
wear properties \\
wear resistance \\
wear test \\
wear tester \\
xray diffraction respectively results show
}{ % CAPTION
Concept Extraction for W2521598419.
}{4} % LABEL

\conceptlst{
Crystallization behaviour of AlSm amorphous alloys. Various Al100xSmx alloys 10 14 have been rapidly solidified by single-roller melt spinning with careful control of the atmosphere in the quenching device. The structural state and subsequent devitrification behaviour of the melt-spun ribbons are found to be particularly sensitive to the quenching conditions. Except for the thinnest ribbons, there are inhomogeneities both through the ribbon thickness and along the length. Both \colorbox{yellow}{fully and partially amorphous ribbons} have been obtained. The crystallization processes of the amorphous phases have been followed by X-ray diffraction, transmission electron microscopy and differential scanning calorimetry DSC. Al92Sm8 shows primary crystallization of Al, followed by the formation of metastable phases; Al90Sm10 transforms polymorphically to a metastable intermetallic; Al88Sm12 and Al86Sm14 display eutectic crystallization into Al and a metastable mixture of compounds. For Al90Sm10, the DSC traces involve several overlapping peaks. This may be the result of transformations occurring in distinct parts of the ribbon with different mechanisms. A kinetic analysis of the crystallization processes has been performed by means of isothermal and non-isothermal DSC experiments. A discussion of the kinetic parameters derived from Kissinger and Avrami analyses is provided.
}{ % Concepts
amorphous alloy \\
avrami analysis \\
crystallization behavior \\
crystallization process \\
devitrification behavior \\
differential scanning calorimetry \\
dsc \\
eutectic crystallization \\
\colorbox{yellow}{fully amorphous ribbon} \\
inhomogeneity \\
isothermal dsc experiment \\
kinetic analysis \\
kissinger analysis \\
melt spinning \\
metastable phase \\
non isothermal dsc experiment \\
\colorbox{yellow}{partially amorphous ribbon} \\
polymorphic transformation \\
primary crystallization \\
quenching device \\
ribbon length \\
ribbon thickness \\
structural state \\
transmission electron microscopy \\
x ray diffraction
}{ % RAKE Concepts
al followed \\
alsm amorphous alloys various alxsmx alloys \\
alsm display eutectic crystallization \\
amorphous phases \\
avrami analyses \\
careful control \\
crystallization behaviour \\
crystallization processes \\
different mechanisms \\
distinct parts \\
dsc traces involve several overlapping peaks \\
kinetic analysis \\
kinetic parameters derived \\
meltspun ribbons \\
metastable intermetallic alsm \\
metastable mixture \\
metastable phases alsm transforms polymorphically \\
nonisothermal dsc experiments \\
partially amorphous ribbons \\
particularly sensitive \\
quenching conditions except \\
quenching device \\
rapidly solidified \\
ribbon thickness \\
singleroller melt spinning \\
structural state \\
subsequent devitrification behaviour \\
thinnest ribbons \\
transformations occurring \\
xray diffraction transmission electron microscopy \\
}{
Concept Extraction for W2047249259.
}{5} % LABEL

\conceptlst{ % ABSTRACT
\colorbox{yellow}{Successive ionic layer adsorption and reaction} SILAR trend for nanocrystalline mercury sulfide thin films growth. Mercury sulfide HgS nanocrystalline thin films have been grown onto amorphous glass substrate by successive ionic layer adsorption and reaction SILAR trend at room temperature 27C. The optimized preparative parameters including ion concentration, number of immersion cycles, and pH of the solution are used for fine nanocrystalline film growth. A further study has been made for the structural, surface morphological, optical and electrical properties of the films by using X-ray diffraction XRD, scanning electron microscopy SEM, transmission electron microscopy TEM, optical absorption and dc two point probe method. The as-deposited grown HgS nanocrystalline films exhibited cubic phase, with optical band gap Eg of 2.0eV and electrical resistivity of the order of 103cm. SEM and TEM images confirmed films of smooth surface morphology and nanocrystaline in nature with fine crystallites of 2030nm diameter, respectively.
}{ % CONCEPTS
amorphous glass substrate \\
cubic phase \\
dc two point probe method \\
electrical property \\
electrical resistivity \\
mercury sulfide \\
nanocrystaline mercury sulfide thin films \\
nanocrystaline mercury sulfide thin films growth \\
optical band gap \\
optical property \\
scanning electron microscopy \\
sem \\
\colorbox{yellow}{successive ionic layer adsorption and reaction} \\
successive ionic layer adsorption and reaction trend \\
transmission electron microscopy \\
x ray diffraction
}{ % RAKE CONCEPTS
dc two point probe method \\
diameter respectively \\
electrical properties \\
electrical resistivity \\
fine crystallites \\
fine nanocrystalline film growth \\
grown onto amorphous glass substrate \\
immersion cycles \\
optical band gap eg \\
reaction silar trend \\
room temperature c \\
smooth surface morphology \\
structural surface morphological optical \\
successive ionic layer adsorption \\
tem images confirmed films
}{ % CAPTION
Concept Extraction for W1993498556 (correct interpretation of 'and'; abbreviation not extracted).
}{6} % LABEL
\clearpage

\section{Concept graph properties}
\begin{sitable}[ht!]
    \centering\small
    \caption{25 most frequent concepts with their occurrences and degree in the concept graph.}
    \begin{tabular}{p{0.35\textwidth}C{0.225\textwidth}C{0.225\textwidth}}\toprule
        Concept &  Occurance% in \dots Works
        &  Degree% in Concept Graph
        \\\midrule
                        mechanical property &  22931 &  261082 \\
                          x ray diffraction &  18021 &  221632 \\
               scanning electron microscopy &  12185 &  154809 \\
           transmission electron microscopy &   9462 &  118857 \\
                           tensile strength &   8840 &  105074 \\
                        electron microscopy &   8001 &   99775 \\
                                        SiC &   6402 &   68771 \\
                                 grain size &   5273 &   60612 \\
                            aluminium alloy &   5075 &   56272 \\
                             grain boundary &   4675 &   53671 \\
                             yield strength &   4202 &   54042 \\
                        plastic deformation &   4151 &   46884 \\
                           optical property &   4125 &   50866 \\
                                       TiO2 &   4085 &   49291 \\
                                      Al2O3 &   3988 &   47438 \\
               scanning electron microscope &   3831 &   46055 \\
                               tensile test &   3814 &   46973 \\
                                       SiO2 &   3749 &   42820 \\
                                        ZnO &   3706 &   45161 \\
                             heat treatment &   3538 &   40340 \\
                        electrical property &   3495 &   40945 \\
                           tensile property &   3362 &   40163 \\
                       corrosion resistance &   3219 &   38498 \\
                                strain rate &   3198 &   36590 \\
                          magnetic property &   3135 &   35066 \\ \bottomrule
    \end{tabular}
    \label{si-tab:frequent-concepts}
\end{sitable}

\begin{sifigure}[ht!]
   \centering\small
   \includegraphics{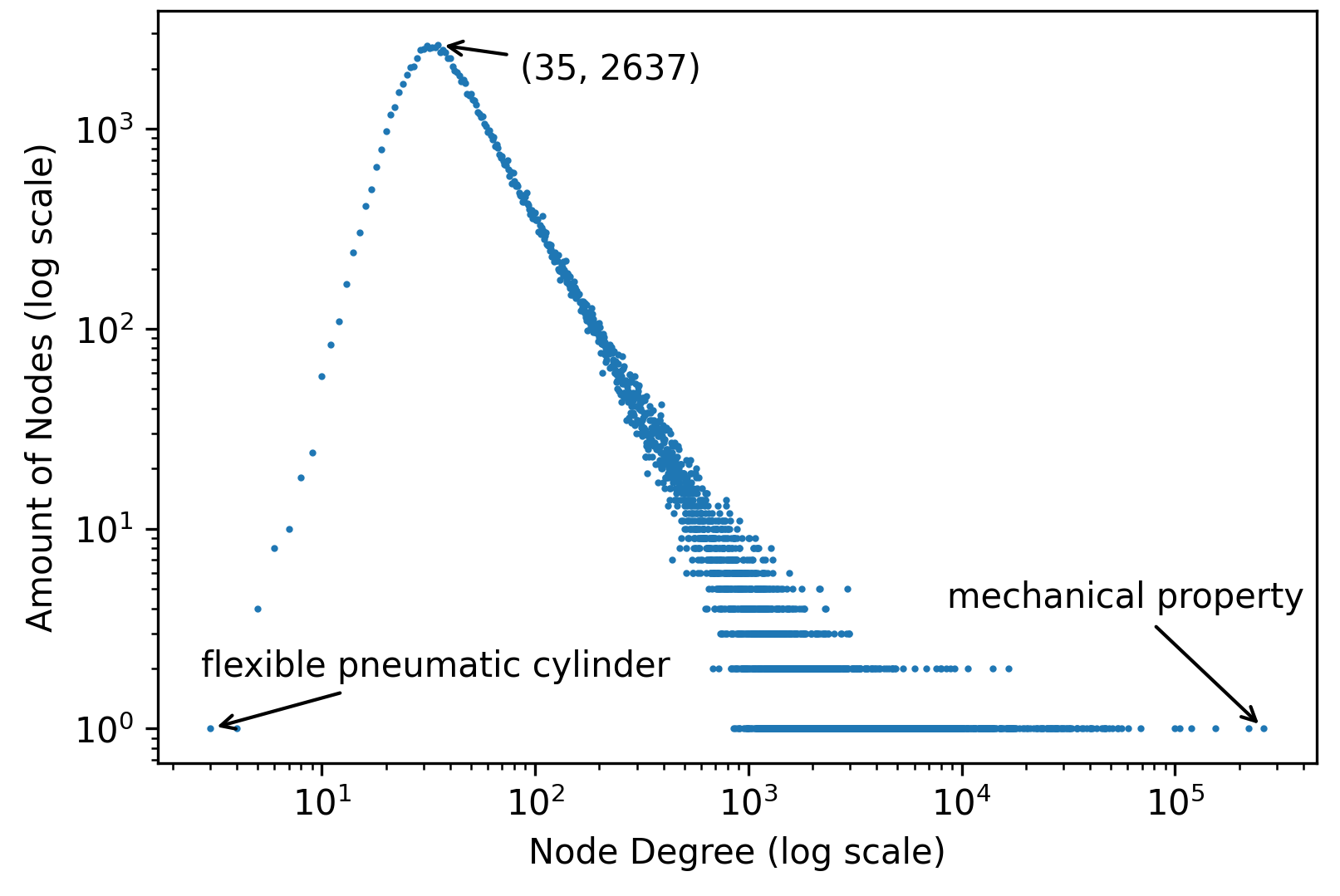}
   \caption{Distribution of node degrees. The node degree of 35 is the most common, occurring 2637 times. The concept with the highest degree is 'mechanical property'. The concept with the lowest degree is 'flexible pneumatic cylinder'. Both axes are in log scale.}
   \label{si-fig:degree-distribution}
\end{sifigure}

\begin{sifigure}[ht!]
   \centering\small
   \includegraphics[scale=0.5]{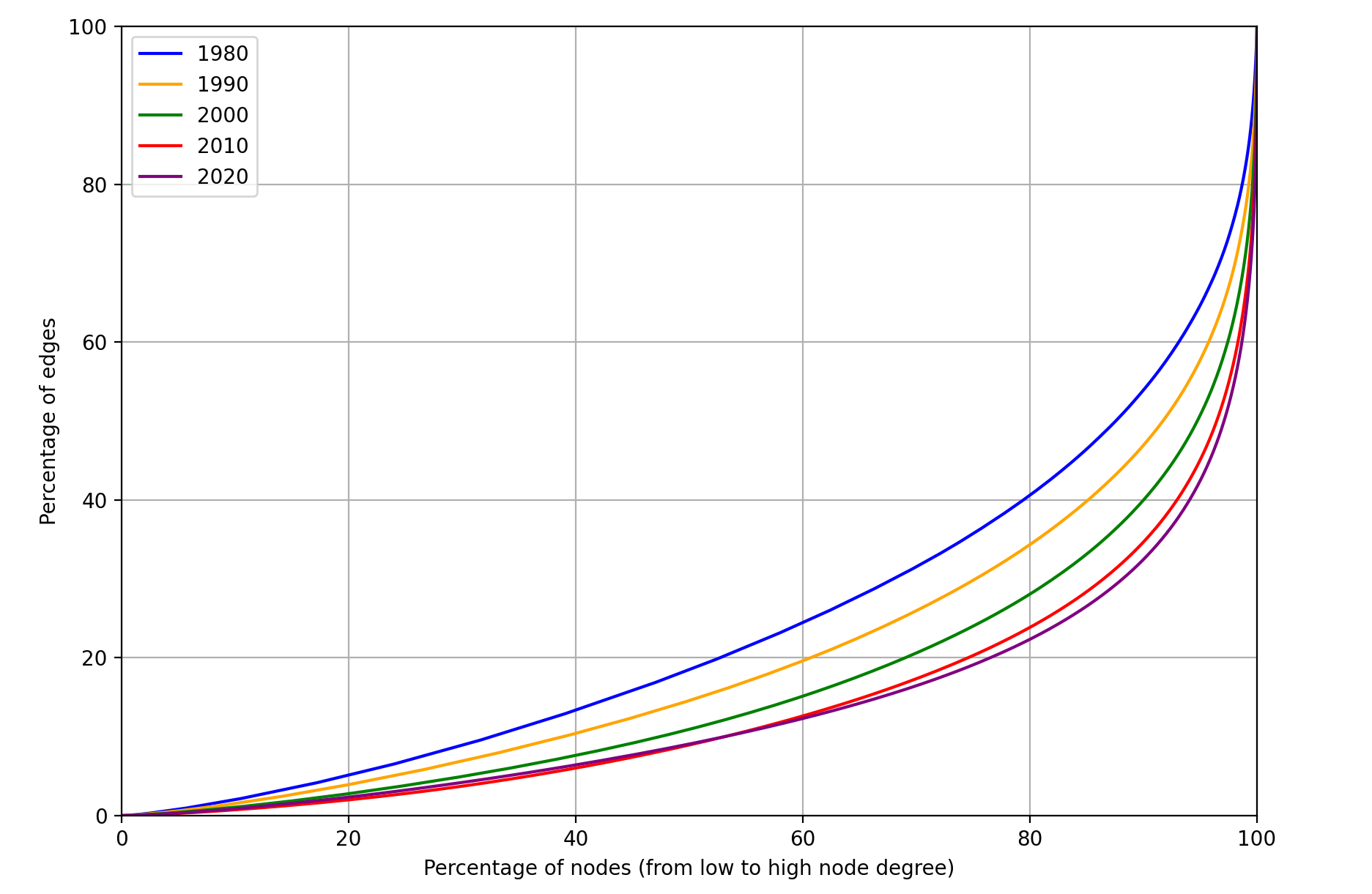}
   \caption{Concept Centralization over different years. The graph illustrates the cumulative percentage of edges connected to nodes when sorted by their degree. A steep curve indicates that a small percentage of nodes (highly central nodes) are connected to a large percentage of edges. The years 1980, 1990, 2000, 2010, and 2020 are compared to show how concept centralization increases over time in our Concept Graph.}
   \label{si-fig:concept-centralization}
\end{sifigure}

% concept embeddings

 \begin{sitable}
    \centering\small
    \caption{5 exemplary concepts and their respective 5  nearest neighbors}   
    \begin{tabular}{p{0.3\textwidth}p{0.4\textwidth}p{0.1\textwidth}}\toprule
    Concept & Neighbor & Distance \\ \midrule
    \multirow{5}{*}{compound semiconductor} & semiconductor compound & 2.798 \\
    & compound semiconductor material & 3.22  \\ 
    & semiconductor material          & 3.457 \\ 
    & semiconducting compound         & 3.584 \\ 
    & semiconductor crystal           & 3.658 \\ \midrule
    \multirow{5}{*}{pyrocarbon matrix} & pyrocarbon layer & 7.07 \\
    & resin carbon matrix     & 8.072 \\
    & pyrolytic carbon matrix & 8.115 \\
    & matrix carbon           & 8.115 \\
    & carbon matrix           & 8.202 \\ \midrule
    \multirow{5}{*}{pure titanium sheet} & pure titanium & 5.801 \\
    & titanium sheet                   & 5.82  \\
    & pure titanium plate              & 6.192 \\
    & commercially pure titanium sheet & 6.594 \\
    & pure titanium surface            & 6.914 \\ \midrule
    \multirow{5}{*}{wear resistant coating} & coating property & 1.428 \\
    & corrosion resistant coating & 1.465 \\
    & coating wear resistance     & 1.663 \\
    & coating protection          & 1.73  \\
    & coating characteristic      & 1.774 \\ \midrule
    \multirow{5}{*}{304l stainless steel} & aisi 304l stainless steel & 4.861 \\
    & 316l stainless steel      & 5.294 \\
    & 304l steel                & 5.323 \\
    & 308l stainless steel      & 5.742 \\
    & type 304l stainless steel & 5.834 \\ \bottomrule
    \end{tabular}
    \label{si-tab:knn}
 \end{sitable}

\clearpage

\section{Link prediction}\label{si-sec:link-pred}
\subsubsection*{Test set construction}
\label{si-sec:test-set-construction}
To get a clear picture of our models' performances, we do not perform oversampling for the positives during test set construction. This is necessary to analyze AUROC, precision and recall in a setting as close to real-world applications as possible. We approach test set construction by randomly sampling node pairs that are not connected until the end of 2019 and simply storing the label by looking three years ahead in the graph. The resulting 2,000,000 node pairs, therefore, mirror the real distribution of positives. Of these 307 (0.015\%) positive examples, 87 links were formed in the year 2020, 123 in 2021, and 97 in 2022.

\subsubsection*{Evaluation against Science4Cast}
\label{si-sec:science4cast}

We benchmark our Baseline model against the Science4Cast benchmark described in \cite{krenn_predicting_2022}. First, we train the model on data up until 2014 to predict links emerging in 2015, 2016 and 2017. Second, we evaluate on the node pairs contained in `SemanticGraph\_delta\_3\_cutoff\_0\_minedge\_1.pkl'. Surprisingly, our Baseline model is not only competitive but would have scored the second place in the Science4Cast competition with an AUROC of $0.9088$. \autoref{si-tab:s4c-performance} presents an exhaustive overview of precision, recall, F1 score and the raw data at different thresholds.

\begin{sitable}
    \centering\small
    \caption{Performance metrics of our Baseline model at different thresholds evaluated against the Science4Cast benchmark on `SemanticGraph\_delta\_3\_cutoff\_0\_minedge\_1.pkl' \cite{krenn_predicting_2022}.}
    \begin{tabular}{c@{\hspace{10pt}}c@{\hspace{10pt}}c@{\hspace{10pt}}c@{\hspace{10pt}}c@{\hspace{10pt}}c@{\hspace{10pt}}c@{\hspace{10pt}}c@{\hspace{10pt}}c}\toprule
        Threshold & AUC & Precision & Recall & F1-score & TN & FP & FN & TP \\\midrule
        0.50 & 0.9088 & 0.0818 & 0.5170 & 0.1413 & 9663029 & 287423 & 23934 & 25614 \\
        0.55 & 0.9088 & 0.0952 & 0.4705 & 0.1583 & 9728873 & 221579 & 26238 & 23310 \\
        0.60 & 0.9088 & 0.1101 & 0.4266 & 0.1751 & 9779674 & 170778 & 28411 & 21137 \\
        0.65 & 0.9088 & 0.1267 & 0.3837 & 0.1905 & 9819397 & 131055 & 30538 & 19010 \\
        0.70 & 0.9088 & 0.1458 & 0.3410 & 0.2042 & 9851432 & 99020  & 32653 & 16895 \\
        0.75 & 0.9088 & 0.1693 & 0.2950 & 0.2152 & 9878748 & 71704  & 34930 & 14618 \\
        0.80 & 0.9088 & 0.2027 & 0.2497 & 0.2238 & 9901799 & 48653  & 37175 & 12373 \\
        0.85 & 0.9088 & 0.2474 & 0.1975 & 0.2196 & 9920681 & 29771  & 39763 & 9785  \\
        0.90 & 0.9088 & 0.3225 & 0.1332 & 0.1885 & 9936587 & 13865  & 42948 & 6600  \\
        0.95 & 0.9088 & 0.4881 & 0.0568 & 0.1017 & 9947501 & 2951   & 46734 & 2814  \\\bottomrule
    \end{tabular}
    \label{si-tab:s4c-performance}
\end{sitable}

\subsubsection*{Analysis of Ensemble Performance Across Feature Domains}

We investigated the impact of model blending on AUC performance, i.e. whether performance gains arise from combining distinct feature modalities or solely from the averaging effect of ensembling. While blends of models with complementary information (embeddings mixed with topological models, Panels 1 and 2 of \autoref{si-fig:auc-hybrid}) achieve substantial performance boosts, blends within the same feature domain (Panel 3) or model type (Panel 4) conversely result in only marginal improvements. This demonstrates that the observed gains are driven by the synergy of non-redundant features rather than simple ensemble averaging.

\begin{sifigure}[h!]
    \centering\small
    \includegraphics[width=\textwidth]{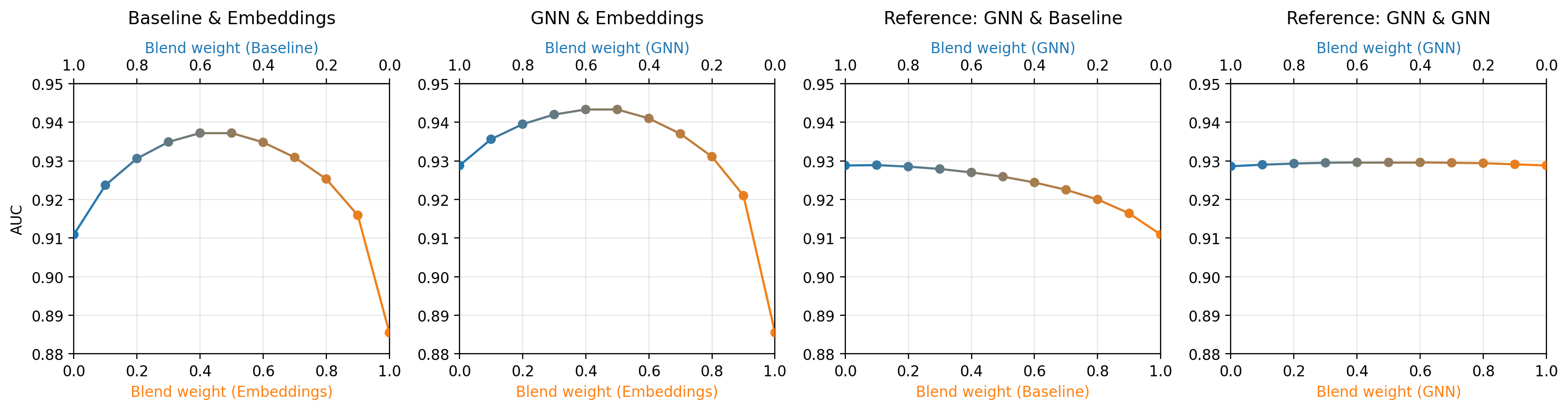}
    \caption{AUC performance for four hybrid models as a function of the blend weights for different combinations of models.}
    \label{si-fig:auc-hybrid}
\end{sifigure}

\subsubsection*{Calibration of the models' predictions}
\begin{sitable}[ht]
    \centering\small
    \caption{Calibration of our models' predictions measured with Brier-score.}
    \begin{tabular}{p{.375\linewidth}p{.125\linewidth}}
        \toprule
        Model & Brier Score \\ \midrule 
        Baseline & 0.0521 \\
        GNN & 0.0953 \\
        Concept Embeddings (MatSciBERT) & 0.0535 \\
        Concept Embeddings (BERT) & 0.0596 \\
        Combination of features & 0.05019 \\
        Mixture of MLP and Embeddings & 0.0438 \\
        Mixture of GNN and Embeddings & 0.0438 \\
        Pure Text Baseline & 0.1044 \\
        \bottomrule
    \end{tabular}
    \label{si-tab:brier-scores}
\end{sitable}

\newpage
\subsubsection*{Precision and recall}

\begin{sifigure}[h!]
    \centering\small
    \includegraphics[width=\textwidth]{img/pr-curves-all.png}
    \caption{Precision-recall (PR) curves for our link prediction models on the test set ($T_{test} = [2020, 2022]$) with a zoom on the y-axis. Markers highlight the performances at a threshold of 0.5. The left panel shows the PR curve for all data points. The central and right panel display the respective PR curves for $d_{\text{prev}} = 2$ and $d_{\text{prev}} = 3$. Best result is in bold.}
\end{sifigure}

\begin{sifigure}[h!]
    \centering\small
    \includegraphics[width=\textwidth]{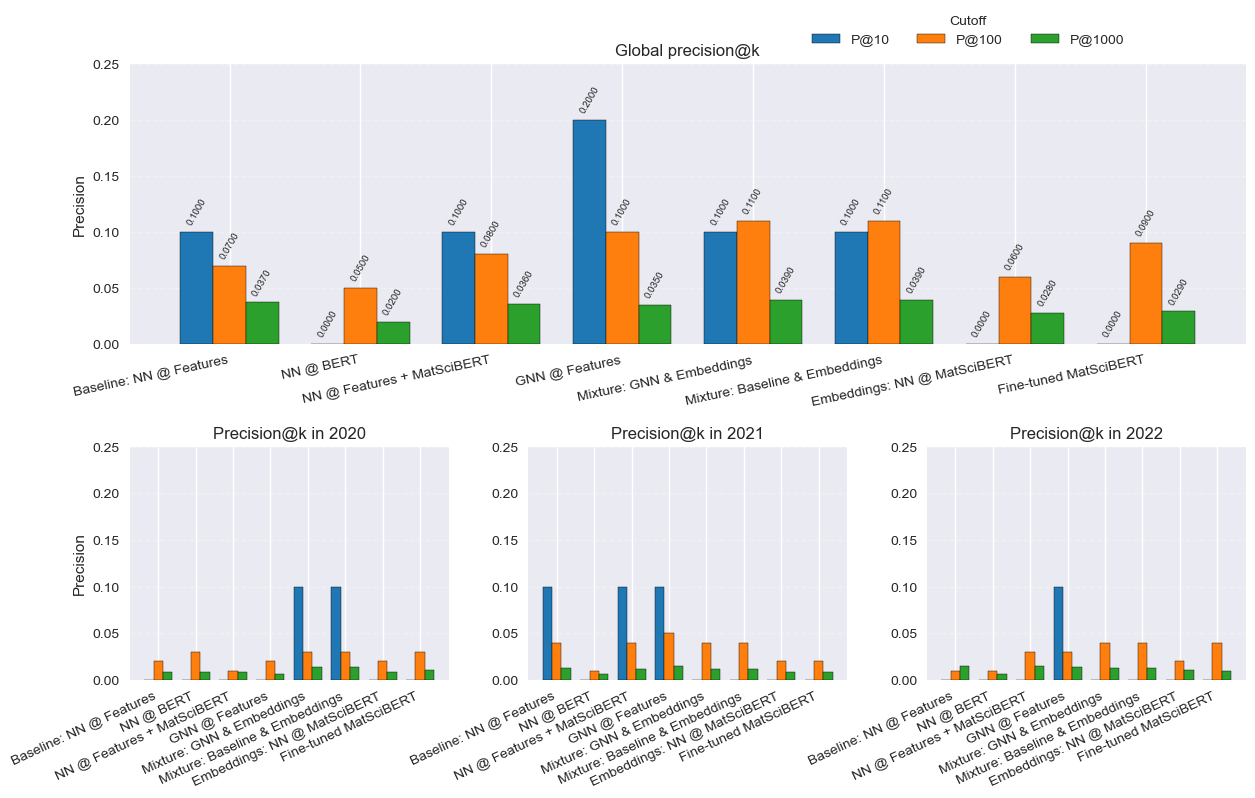}
    \caption{Precision@k for $k \in \{10, 100, 1000\}$ for our link prediction models for all data points, also broken down by year (2020, 2021, 2022).}
    \label{si-fig:extract-embeddings-1}
\end{sifigure}

\begin{sifigure}
    \centering\small
    \includegraphics[width=\textwidth]{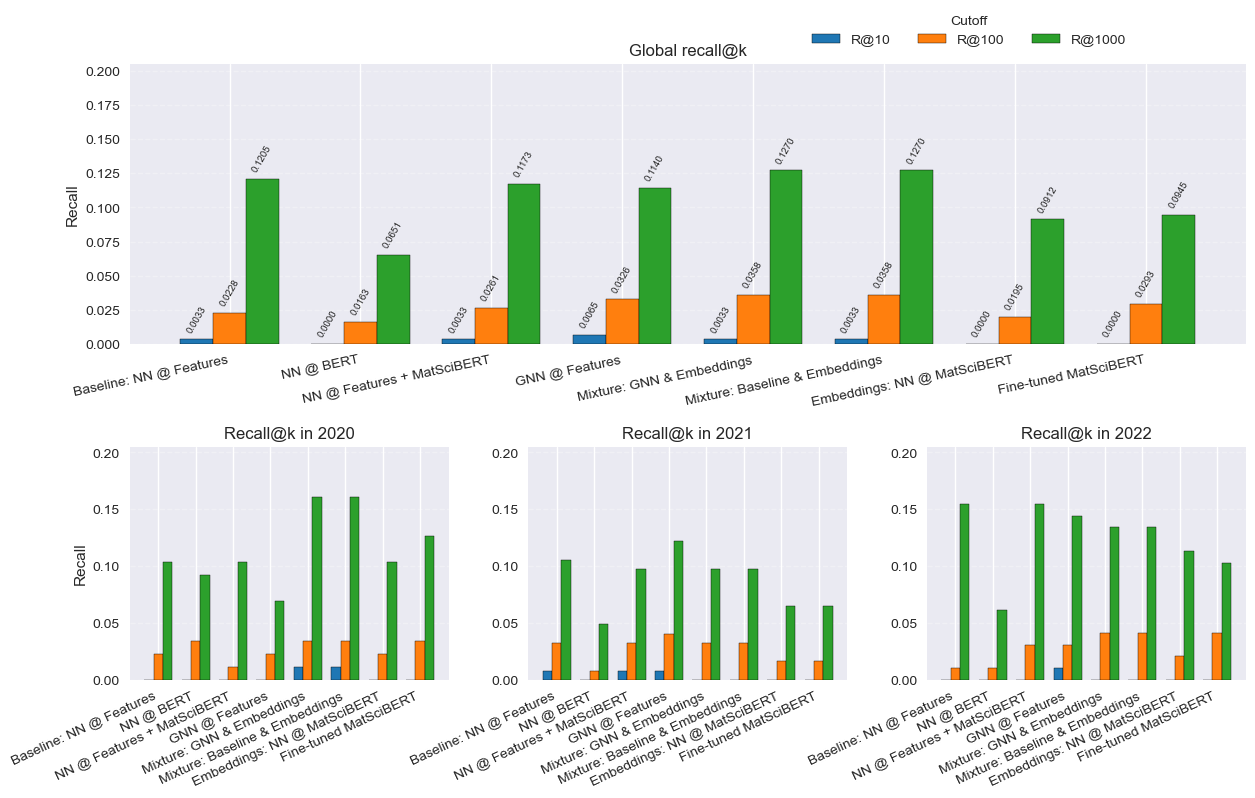}
    \caption{Recall@k for $k \in \{10, 100, 1000\}$ for our link prediction models for all data points, also broken down by year (2020, 2021, 2022).}
    \label{si-fig:extract-embeddings-3}
\end{sifigure}

\begin{sifigure}
    \centering\small
    \includegraphics[width=\textwidth]{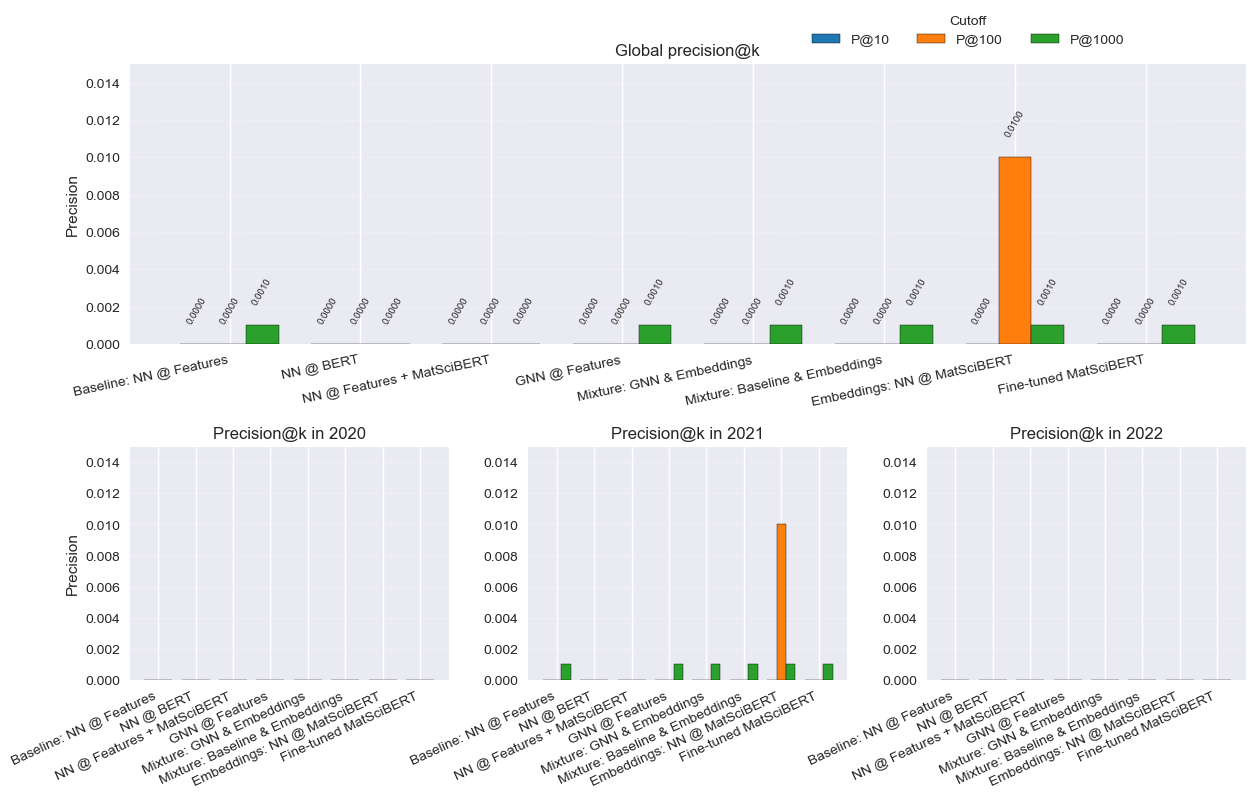}
    \caption{Precision@k for $k \in \{10, 100, 1000\}$ for our link prediction models for $d_\text{prev} = 3$, also broken down by year (2020, 2021, 2022).}
    \label{si-fig:extract-embeddings-2}
\end{sifigure}

\begin{sifigure}
    \centering\small
    \includegraphics[width=\textwidth]{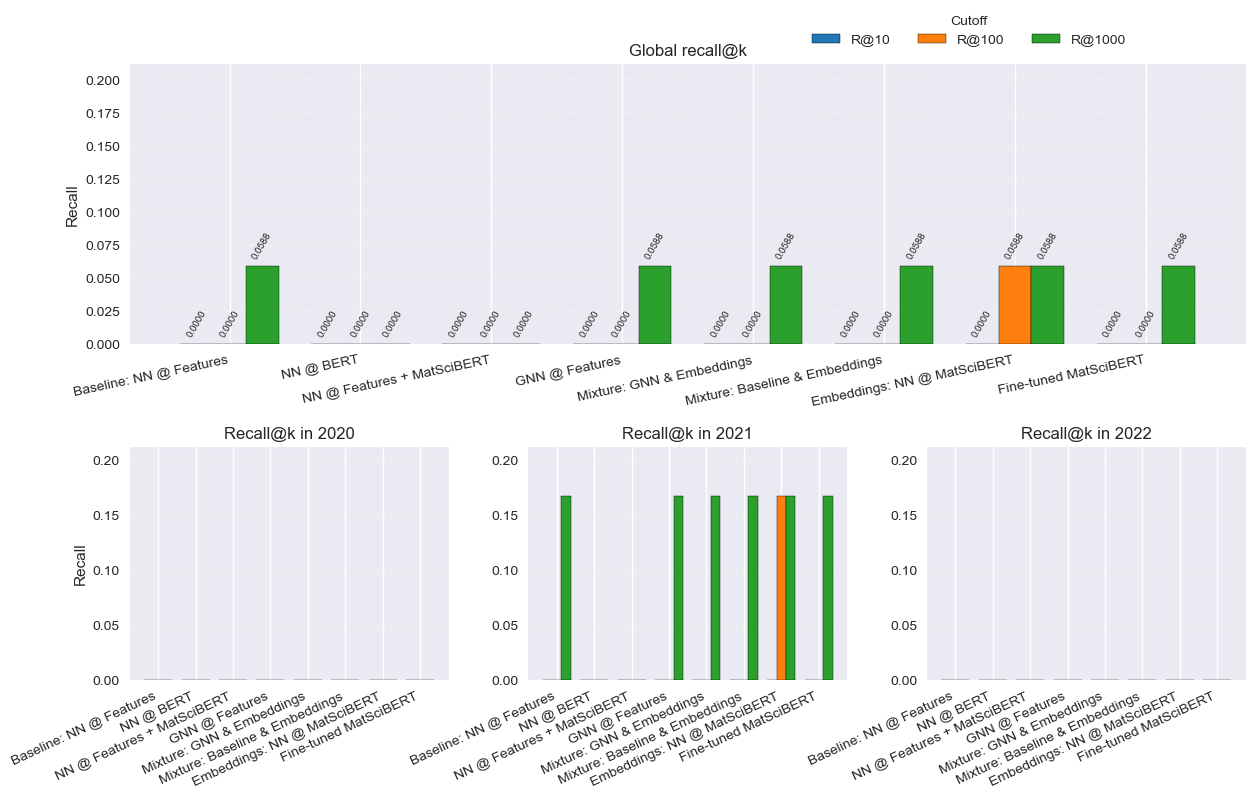}
    \caption{Recall@k for $k \in \{10, 100, 1000\}$ for our link prediction models for $d_\text{prev} = 3$, also broken down by year (2020, 2021, 2022).}
    \label{si-fig:extract-embeddings-4}
\end{sifigure}

\clearpage

\section{Report generation}\label{si-sec:report}
\begin{sifigure}[h!]
    \centering\small
    \includegraphics[width=0.4\textwidth]{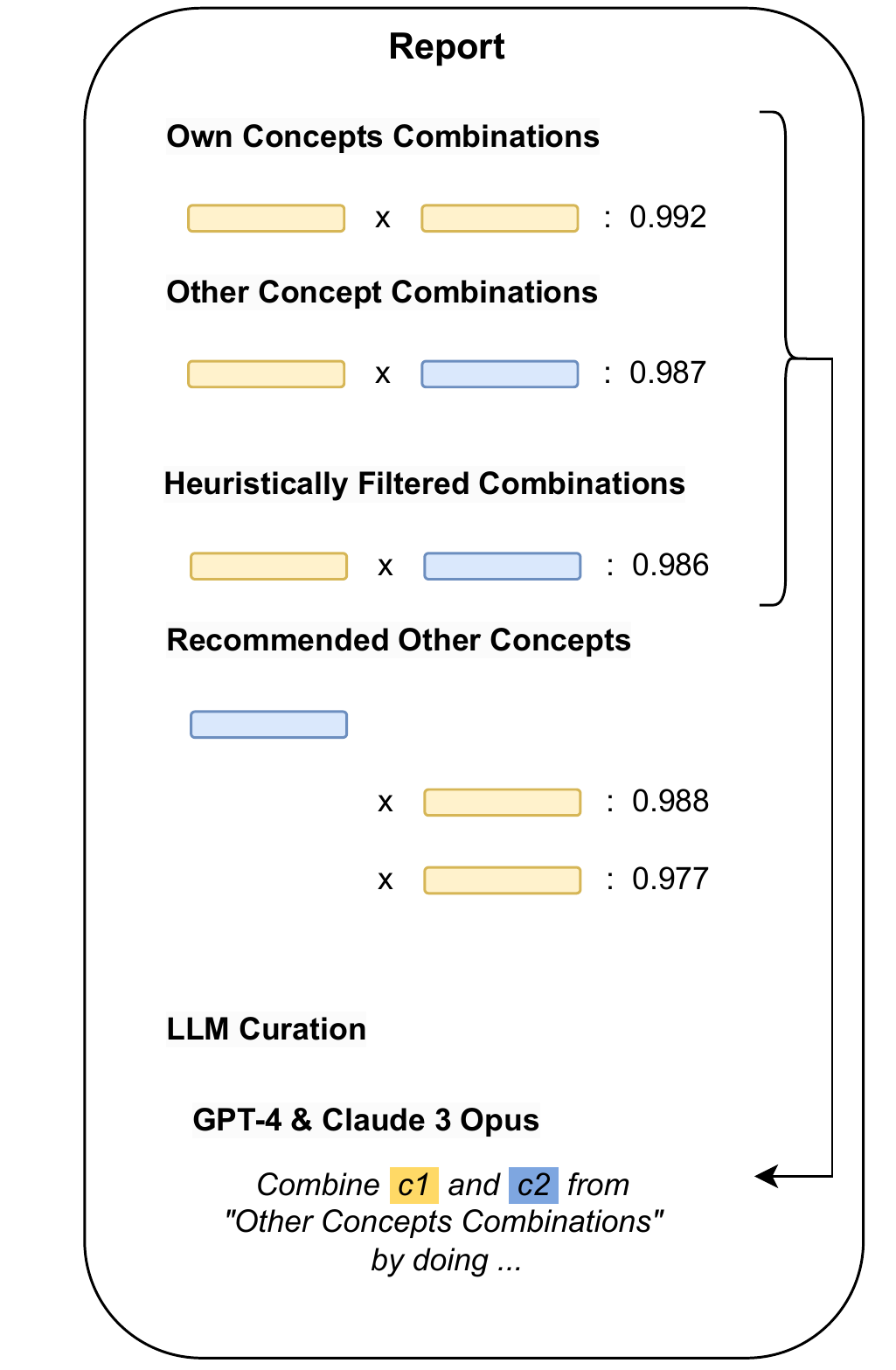}
    \caption{Schematic representation of the report structure. 'Own' ($C_{\text{own}}$) and 'Other' ($C_{\text{other}}$) concepts are respectively represented in yellow and blue.}
    \label{si-fig:report-structure}
\end{sifigure}

\paragraph{Own Concepts Combinations.} We generate combinations by combining the concepts that describe a researcher $C_{\text{own}}$ with themselves: 
\begin{align*}
S_{\text{own}\times\text{own}} := \{(c_1,c_2) \in C_{\text{own}} \times C_{\text{own}} | (c_1, c_2) \notin E(G)\}\quad .
\end{align*}
These combinations are then ranked, and the best 25 are included. The benefit of this section is that researchers are directly able to judge on the usefulness of these suggestions as both concepts are within their expertise. The size of $C_{\text{own}}$ varies between researchers, we observe values roughly between 100 and 400.

\paragraph{Other Concept Combinations.} We generate more diverse combinations by combining $C_{\text{own}}$ with all remaining concepts: 
\begin{align*}
S_{\text{own}\times\text{other}} := \{(c_1,c_2) \in C_{\text{\text{own}}} \times C_{\text{{\text{other}}}} | (c_1, c_2) \notin E(G)\} 
\end{align*}
where $C_{{\text{other}}} := C_{\text{known}} \setminus C_{\text{own}}$. These combinations are then ranked and the best 25 are included. Note that $|Combinations_{\text{own},{\text{other}}}| \gg |Combinations_{\text{own},\text{own}}|$ as the size of $C_{\text{known}}$ is 137,000. The idea of this section is to offer more unbound suggestions than in `Own Concepts Combinations'.

\paragraph{Filtered Other Concept Combinations. } We employ two heuristics to filter through the large set of 
$S_{\text{own}\times\text{other}}$
\begin{align*}
S^{\text{filtered}}_{\text{own}\times\text{other}} := \{ &(c_{\text{own}},c_{{\text{other}}}) \in Combinations_{\text{own},{\text{other}}} \\| &\deg(c_{{\text{other}}}) <= 100 \And 3 <= \text{dist}(c_{\text{own}},c_{{\text{other}}} <= 6) \} \quad .
\end{align*} To avoid unspecific suggestions, i.e. broad terms like `materials science', we exclude concepts based on their \textit{degree} in the concept graph $G$. To prevent obvious, e.g. synonyms, and too far-fetched concepts, we integrate semantic knowledge by using the euclidean distance in the map of the materials science (\textbf{Figure~1}) that is a two dimensional projection of the obtained concept embeddings. Note that the values 3 and 6 were arbitrarily chosen after examining the map of materials science.

\paragraph{Other Concepts x Many Own Concepts.} In this section, we suggest single concepts that have highly scored connections with many own concepts $C_{\text{own}}$: 
\begin{align*}
S_{\text{(many own)}\times\text{other}} 
&:= \{(c_{{\text{other}}}, C_{\text{own},\text{matching}}) \in C_{{\text{other}}} \times \{ C_{\text{own}} \times ... \times C_{\text{own}} \} | C_{\text{own},\text{matching}} \\ 
&:= \{ c \in C_{\text{own}} | \text{score}(c_{{\text{other}}},c) > p \} \}
\end{align*}
where p can be set freely as threshold. The resulting set is sorted descendingly according to the length of $C_{\text{own},\text{matching}}$ and the top 20 are included. The idea is to take the researcher's profile into account as a whole and to suggest concepts that can therefore be connected more easily to one's own research.

\paragraph{LLM Curation}
We are interested in the capabilities of LLMs to assist humans by pre-selecting from a broader list of concept combinations. For this, we query two LLMs, GPT-4 and Claude 3 Opus (access May/June 2024), to select three items from a list of combinations and to write an additional paragraph on why this combination is promising. Three combinations were selected from $S_{\text{own}\times\text{own}} \cup S_{\text{own}\times\text{other}}$ together with three additional combinations from $S^{\text{filtered}}_{\text{own}\times\text{other}}$, resulting in a total of six suggested combinations per LLM per report, and the two prompts are given in \multiautoref{si-fig:prompt_own_own, si-fig:prompt_own_other}.

No paragraphs were created for combinations from section $S_{\text{(many own)}\times\text{other}}$ as we expected that having the LLM generate a full paragraph combining multiple concepts would be harder to digest and analyze within the limited interview time.
Note that this report section does not contain new suggestions but rather selects and elaborates on combinations already present in the report. Apart from highlighting particular suggestions, the LLM approach helps to bridge the gap between two concepts by providing additional information on how they could be combined. This is especially helpful for $S_{\text{own}\times\text{other}}$ or $S^{\text{filtered}}_{\text{own}\times\text{other}}$ where the second concept might be unfamiliar to a researcher.

As our prompt to select 3 combinations of concepts from section 3 (\autoref{si-fig:prompt_own_other}) explicitly asked the LLM to choose “more exotic” combinations, we analyzed if this resulted in a bias towards far-distance, cross-field pairs. For Sections 1 and 2, the mean node degree of the selected pairs was slightly lower (9,712) than for the non-selected pairs (11718). While all of the selected pairs were previously connected through only one additional node ($d_{\text{prev}}=2$), 2 of the non-selected pairs showed a larger previous node distance of $d_{\text{prev}}=3$. For section 3, mean node degrees of 1,582 and 4,694 were found for the selected and non-selected pairs, respectively. Here, 4 of the selected and 3 of the non-selected pairs showed a previous node distance of $d_{\text{prev}}=3$. While the lower node degrees as well as the slightly higher number of more distant concepts selected from section 3 could result from the wording in the prompt, the small sample size prevents any definitive conclusions.

\begin{sitable}[ht!]
    \centering\small
    \caption{LLM suggestion vs. "Is Interesting" (labeled as C) confusion matrix.}
    \begin{tabular}{C{0.15\textwidth}C{0.06\textwidth}C{0.06\textwidth}C{0.06\textwidth}C{0.06\textwidth}}
    \toprule
        \multicolumn{1}{c}{} & & \multicolumn{3}{C{0.18\textwidth}}{LLM suggestion}\\
        \multicolumn{1}{c}{} & & Y & N & $\sum$ \\\midrule
        \multirow{3}{*}{"Is Interesting"} & Y & 24 &  37 & 61 \\
        & N &  29 &  24 & 205\\
        & $\sum$ & 53 & 213 & 266\\\bottomrule
    \end{tabular}
    \label{si-table:llm-interesting-sugg-confusion-matrix}
\end{sitable}

%\paragraph{Prompts.} We present the two prompts we used to select interesting combinations and generate abstracts on how the presented combinations could be realized. The prompt below was used to select 3 concept combinations from report sections 1 and 2:

\begin{sifigure}[bht]
\centering
\begin{lstlisting}[xleftmargin=0pt, framexleftmargin=0pt]
You are an advanced AI agent who has expertise in the realm of the materials sciences. Your goal is to assist researchers looking for promising new concept combinations that will --- once investigated --- advance mankind.

We prepared a report containing several concept combinations suggested by an ML model we trained on historical data. This is an overview of the attached PDF:

1.1 Possible new combinations of the concepts we identified for a particular scientist (which, to our knowledge, have not yet been combined)
1.2 Possible new combinations of the concepts we identified for a particular scientist with other existing concepts that, to our knowledge, the scientist has not yet worked on (at least not according to the abstract data of the last years)

Which 3 combinations (across both categories) do you think (including all material science knowledge you have) are the most promising? The order of the combinations doesn't indicate significance. Write a concise paragraph for each of those suggestions that explains why you chose this combination. Each suggested combination must be contained in the report. Don't invent new combinations that are not present.

Just give me the 3 combinations and the 3 paragraphs.
\end{lstlisting}
\caption{Prompt for the selection of 3 concept combinations from report sections 1 ($S_{\text{own}\times\text{own}}$) and 2 ($S_{\text{own}\times\text{other}}$).}
\label{si-fig:prompt_own_own}
\end{sifigure}

%The prompt below was used to select 3 concept combinations from report section 4:

\begin{sifigure}[ht!]
\begin{lstlisting}[xleftmargin=0pt, framexleftmargin=0pt]
You are an advanced AI agent who has expertise in the realm of the materials sciences. Your goal is to assist pairs of researchers looking for promising areas to combine their research which will - once investigated - advance mankind.

We prepared a report containing several concept combinations suggested by an ML model we trained on historical data. This is an overview of the attached PDF:

1. Possible new concept combinations that are more exotic. 

For each combination, imagine two researchers proficient in one of the two fields/concepts that want to collaborate.

Which of these 3 of these more exotic combinations do you think (including all material science knowledge you have) are the most promising?

The order of the combinations doesn't indicate significance. Write a concise paragraph for each of those suggestions that explains why you chose this combination, and extend the paragraph by thinking about how the two scientists would collaborate and develop ideas and pathways to establish this cooperation.

Each suggested combination must be contained in the report. Don't invent new combinations that are not present.

Just give me the 3 combinations and the 3 paragraphs.
\end{lstlisting}
\caption{Prompt for the selection of 3 concept combinations from report section 3 ($S^{\text{filtered}}_{\text{own}\times\text{other}}$).}
\label{si-fig:prompt_own_other}
\end{sifigure}

%Example output:

\begin{sifigure}[ht!]
\begin{lstlisting}[xleftmargin=0pt, framexleftmargin=0pt]
Nanoporous Structure and High Entropy Alloy

Nanoporous structures combined with high entropy alloys (HEAs) represent an intriguing area of study due to the unique properties HEAs offer, such as exceptional mechanical strength, corrosion resistance, and thermal stability. The introduction of a nanoporous architecture can potentially enhance these properties by introducing lightweight characteristics and increasing surface area, which could be particularly beneficial for applications in catalysis and energy storage. Combining these two concepts could lead to the development of new, optimized materials with tailored properties for specific industrial applications.

In Situ High Resolution Transmission Electron Microscopy and Mechanical Performance

The application of in situ high-resolution transmission electron microscopy (HRTEM) to study the mechanical performance of materials offers a promising research direction. This technique allows real-time imaging of material deformation, phase transformations, and dislocation dynamics at the atomic scale during mechanical testing. Understanding these microscopic events in great detail will enable the design of materials with superior mechanical properties tailored to specific applications, such clarification provides deeper insights into materials behavior under stress and can lead to the development of materials with enhanced durability and performance.

Epitaxial Strain and Ultraviolet Visible Spectroscopy

Studying the effects of epitaxial strain using ultraviolet-visible spectroscopy could open new pathways in the synthesis and characterization of thin films and nanostructures. Epitaxial strain has been known to significantly alter the electronic and optical properties of materials. By applying UV-Vis spectroscopy, researchers can gain valuable insights into the changes in band structure and electronic transitions caused by strain. This combination could be particularly valuable in the development of advanced optical devices, including lasers, LEDs, and photovoltaic cells, where precise control over electronic properties is crucial.
\end{lstlisting}
\caption{Example of an LLM output containing three concept combinations and the corresponding abstracts}
\end{sifigure}
\clearpage

\section{Rated Concept Combinations}
\subsubsection*{Selection of the interviewees}
The selection of 10 interviewed researchers included only PIs, mostly professors at universities, from a diverse set of subdisciplines of materials science, some of which, but not all, had previous research experience with machine learning models.

\subsubsection*{Analysis of the rated concepts per researcher}
\begin{sifigure}[h!]
    \centering\small
    \includegraphics[width=\textwidth]{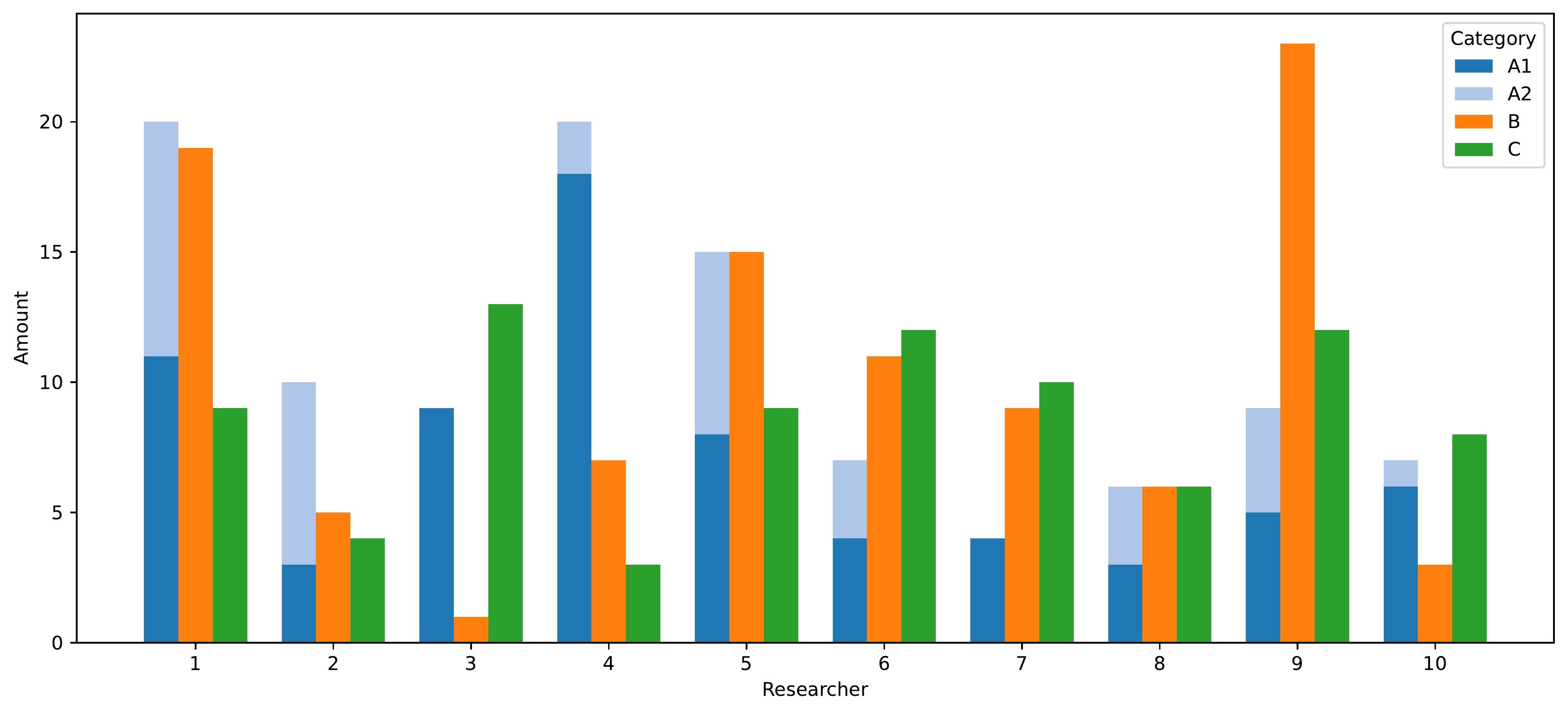}
    \caption{Overview of the concept categories per interview. A1 and A2 are grouped together. Note that the results from researcher 3 were obtained by a commentary on the report, not by an interview.}
    \label{si-fig:overview-concepts-per-researcher}
\end{sifigure}

\begin{sitable}[h!]
    \centering\small
    \caption{Category counts per researcher with per-participant variance.}
    \begin{tabular}{p{0.15\textwidth}p{0.08\textwidth}p{0.08\textwidth}p{0.08\textwidth}p{0.08\textwidth}p{0.08\textwidth}p{0.12\textwidth}}\toprule
    Researcher & A1 & A2 & B & C & D & Variance \\ \midrule
    1  & 11 & 9 & 19 & 9 & 0 & 36.64 \\
    2  & 3  & 7 & 5  & 4 & 0 & 5.36  \\
    3  & 9  & 0 & 1  & 12& 1 & 24.24 \\
    4  & 18 & 2 & 7  & 3 & 0 & 41.20 \\
    5  & 8  & 7 & 15 & 9 & 0 & 22.96 \\
    6  & 4  & 3 & 11 & 11& 1 & 17.60 \\
    7  & 4  & 0 & 9  & 8 & 2 & 11.84 \\
    8  & 3  & 3 & 6  & 6 & 0 & 5.04  \\
    9  & 5  & 4 & 23 & 7 & 5 & 51.36 \\
    10 & 6  & 1 & 3  & 8 & 0 & 9.04  \\ \bottomrule
    \end{tabular}
    \label{si-tab:overview-concepts-per-researcher}
\end{sitable}

\newpage
\subsubsection*{Examples of rated concept combinations}
\begin{sitable}[h!]
    \centering\small
    \caption{Concept combinations ranked A1 and A2 during the interviews (25 examples per category)}
    \begin{tabular}{C{0.15\textwidth}p{0.375\textwidth}p{0.375\textwidth}C{0.1\textwidth}}            
    \toprule
    Category & \multicolumn{1}{c}{Concept A} &  \multicolumn{1}{c}{Concept B} & Section\textsuperscript{[a]} \\ \midrule
    \multirow{25}{*}{A1} & nanoparticle dispersion & quantum yield & 3 \\
    & ferroelectric polarization & layered material & 3 \\
    & in plane polarization & microstructure feature & 3 \\
    & nanoparticle dispersion & post heat treatment & 3 \\
    & microstructure characterization & molecular architecture & 4 \\
    & energy dispersive x ray spectroscopy & domain structure & 4 \\
    & ferroelectric polarization & thermal treatment & 1 \\
    & grain texture & power conversion efficiency & 1 \\
    & out of plane polarization & optoelectronic application & 1 \\
    & polymer fullerene solar cell & kelvin probe force microscopy & 1 \\
    & thermal treatment & crystal orientation & 1 \\
    & computational material science & corrosion resistance & 2 \\
    & ion extraction & density functional theory calculation & 2 \\
    & solid electrolyte interphase layer & finite element analysis & 2 \\
    & device architecture & CO2 & 2 \\
    & aging process & density functional theory & 1 \\
    & solution concentration & machine learning & 1 \\
    & acceptor material & crystal structure & 1 \\
    & excitation intensity & thermal stability & 1 \\
    & lattice distortion & optical analysis & 3 \\
    & absorption spectroscopy & additive manufacturing technology & 3 \\
    & artificial neural network & micro structural effect, crack initiation & 4 \\
    & crystalline coating & laser powder bed fusion & 2 \\
    & morphological change & composite material & 2 \\
    & carbon enrichment & scanning electron microscope & 2 \\
    
    \midrule
    \multirow{25}{*}{A2} & electrode fabrication & oxygen concentration & 3 \\
    & optical property & high entropy alloy & 2 \\
    & optoelectronic application & mechanical performance & 2 \\
    & bulk heterojunction & thermal annealing & 1 \\
    & circular dichroism & optoelectronic property & 1 \\
    & electrode layer & quantum efficiency & 1 \\
    & external quantum efficiency & thermal treatment & 1 \\
    & fullerene derivative & optoelectronic property & 1 \\
    & high reflectivity & perovskite solar cell & 1 \\
    & pseudocapacitive material & composite material & 2 \\
    & channel doping & x ray diffraction & 1 \\
    & channel doping & x ray photoelectron spectroscopy & 1 \\
    & chargedischarge test & thermal stability & 1 \\
    & industrial application & surface pattern, cell membrane & 4 \\
    & solid electrolyte interphase & structural property & 1 \\
    & void filling & atom probe tomography & 1 \\
    & resistance change & mechanical performance & 2 \\
    & dislocation plasticity & gradient structure & 3 \\
    & photon flux & high density & 3 \\
    & edge defect & carbon nanotube & 2 \\
    & femtosecond laser & density functional theory & 1 \\
    & light irradiation & density functional theory & 1 \\
    & multimodal imaging & photoluminescence spectroscopy & 1 \\
    & solution synthesis & material design & 1 \\
    & three dimensional printing & optical property & 1 \\
    \bottomrule\addlinespace[\belowrulesep]
    \multicolumn{4}{p{\textwidth}}{[a] Report sections: 1 ($S_{\text{own}\times\text{own}}$), 2 ($S_{\text{own}\times\text{other}}$) , 3 ($S^{\text{filtered}}_{\text{own}\times\text{other}}$) 4 ($S_{\text{(many own)}\times\text{other}}$)}
    \end{tabular}
        \label{si-tab:ranked-combinations_a12}
\end{sitable}
\begin{sitable}
    \centering\small
    \caption{Concept combinations ranked A1 and A2 during the interviews (25 examples per category)}
    \begin{tabular}{C{0.15\textwidth}p{0.375\textwidth}p{0.375\textwidth}C{0.1\textwidth}}            
    \toprule
    Category & \multicolumn{1}{c}{Concept A} &  \multicolumn{1}{c}{Concept B} & Section\textsuperscript{[a]} \\ \midrule
    \multirow{26}{*}{B} & photovoltaic application & microscopic analysis & 3 \\
    & tailored property & deformed grain & 3 \\
    & nanoparticle dispersion & energy absorption capacity & 3 \\
    & material library & tension compression asymmetry & 3 \\
    & optical property & mendelevium simulation & 3 \\
    & crystal orientation & diffusion kinetic & 3 \\
    & ferroelectric domain & interface bonding & 3 \\
    & electrostatic repulsion & NS & 3 \\
    & charge carrier transport & hydrophobic surface & 3 \\
    & OH & crystal phase & 4 \\
    & absorbance spectroscopy & tensile test & 2 \\
    & polymer fullerene solar cell & laser powder bed fusion & 2 \\
    & exciton diffusion length & x ray photoelectron spectroscopy & 2 \\
    & cone beam & high entropy alloy & 2 \\
    & cone beam & optoelectronic application & 1 \\
    & hansen solubility parameter & optoelectronic application & 1 \\
    & light element & optical property & 1 \\
    & microfluidic chip & band gap & 1 \\
    & rare earth oxide & electrostatic interaction & 1 \\
    & artificial intelligence & BO & 3 \\
    & \multirow{2}{*}{ultimate tensile strength} & gas evolution, material library, & \multirow{2}{*}{4}\\
    & & sub threshold swing & \\
    & spectroscopic technique & yield strength & 2 \\
    & cycling performance & solar cell & 1 \\
    & lithium ion battery application & high entropy alloy & 1 \\
    & light conversion & mechanical property & 2 \\
    \midrule
    \multirow{28}{*}{C} & charge carrier transport & hydrophobic surface & 3 \\
    & conjugated polymer & elastic recovery & 3 \\
    & \multirow{2}{*}{electron backscatter diffraction} & solar cell fabrication, & \multirow{2}{*}{4} \\
    & & thin film deposition &\\
    & \multirow{2}{*}{machine learning} & interpenetrating network, & \multirow{2}{*}{4} \\
    & & fluorescence quenching & \\
    & \multirow{2}{*}{mechanical performance} & polymer fullerene solar cell, & \multirow{2}{*}{4} \\
    & & ferroelectric property & \\
    & carbon based material & photocatalytic activity & 2 \\
    & charge carrier transport & compressive strength & 2 \\
    & crystal orientation & organic solar cell & 1 \\
    & in plane polarization & organic solar cell & 1 \\
    & mechanochemical reaction & tensile performance & 3 \\
    & delafossite structure & dual phase & 3 \\
    & dislocation accumulation & capacity fade, high entropy & 4 \\
    & conventional ceramic & graphene oxide & 2 \\
    & high entropy alloy & charge separation & 4 \\
    & microstructure evolution & economical advantage & 4 \\
    & semiconductor substrate & microstructure evolution & 2 \\
    & mid infrared & molecular dynamic simulation & 2 \\
    & exciton binding & hcp phase & 3 \\
    & sodium iodide & microstructural defect & 3 \\
    & metal organic framework & good ductility & 4 \\
    & direct laser writing & youngs modulus & 1 \\    
    & graphene oxide & monte carlo simulation & 4 \\
    & oxide transistor & cyclic voltammetry & 2 \\
    & oxygen vacancy & crack tip plasticity & 4 \\
    & stress induced phase transformation & hexagonal boron nitride & 3 \\
    \midrule
    \multirow{4}{*}{D} & power consumption & nanoparticle formation & 3 \\
   & quantum efficiency & additive manufacturing & 2 \\
   & bauschinger effect & ultimate tensile strength & 2 \\
   & moores law & high entropy alloy & 1 \\
   \bottomrule\addlinespace[\belowrulesep]
   \multicolumn{4}{p{\textwidth}}{[a] Report sections: 1 ($S_{\text{own}\times\text{own}}$), 2 ($S_{\text{own}\times\text{other}}$) , 3 ($S^{\text{filtered}}_{\text{own}\times\text{other}}$) 4 ($S_{\text{(many own)}\times\text{other}}$)}
   \label{si-tab:ranked-combinations_bcd}
\end{tabular}
\end{sitable}
\clearpage

\section{Detailed discussion of interesting concept combinations}
Below is a more detailed discussion of the 5 selected examples of new combinations of concepts shown in the results section of the main text.

\subsubsection*{Suggestion: "Conventional ceramic" + "Graphene oxide"}
% Ben Breitung

During the concept prediction phase of the AI model trained on open-access scientific abstracts, an interesting connection between two research areas emerged. The AI identified “conventional ceramics” and “graphene oxide” as a potentially promising combination. While the integration of graphene oxide with ceramics has been explored in numerous studies, these approaches typically rely on different synthesis methods and subsequent mixing. In contrast, the method presented here is both unique and simple, yielding a distinctive material: a nanometer-thin multilayer graphene scaffold coated with an approximately 200  thick iron oxide layer (see \textbf{Extended Data Fig.~2}). This AI-predicted research direction aligned with prior experimental work that had been conducted but never published, highlighting the model’s potential to identify overlooked but valuable material synthesis strategies.

At first glance, the connection between conventional ceramics and graphene oxide is not obvious due to their fundamentally different structures and chemical behaviors. Conventional ceramics are typically ionic solids, composed of various cation-anion combinations (e.g., transition metals as cations and oxides, fluorides, or sulfides as anions). They are known for their rigidity, thermal and chemical stability, and widespread use in electronic and electrochemical applications. In contrast, graphene oxide consists of covalently bonded carbon and oxygen atoms, forming a highly flexible material that exfoliates and decomposes at elevated temperatures, producing carbon oxide species and exfoliated (multilayer)graphene.

The link between these material classes becomes evident when they are combined into a composite system. However, this integration is challenging due to their distinct synthesis methods and reactivity. In previous experiments aimed at developing a high-conductivity electrode material for conversion type Li-ion batteries, a successful combination of these materials was achieved. Graphene oxide was mixed with iron pentacarbonyl (\ce{Fe(CO)5}), a liquid at ambient conditions, and subjected to autoclave treatment at temperatures above \SI{250}{\degreeCelsius}. Under these conditions, \ce{Fe(CO)5} vaporized and decomposed into atomic Fe(0) species in the gas phase at around \SI{200}{\degreeCelsius}. These iron atoms subsequently reacted with the oxygen groups in graphene oxide, forming iron oxide species (\ce{Fe2O3}, \ce{Fe3O4}) while simultaneously reducing graphene oxide to graphene.

The resulting composite featured a highly conductive multilayer graphene scaffold coated with an approximately \SI{200}{\nm} thick layer of iron oxide. This material exhibited high electrochemical capacity and reversibility, making it a strong candidate for conversion-type electrode materials. The improved redox kinetics were attributed to the high electronic conductivity of graphene and the intimate contact with the thin iron oxide layer, enhancing the overall electrochemical performance.

\subsubsection*{Suggestion: "Tensile strain" + "Molecular architecture"}
% Christoph Brabec

Managing tensile strain has been underestimated in the design of stable interfaces for thin film soft matter' solar cells. In inorganic crystalline semiconductors, strain / stress management has been understood as an important design criterion to prevent the formation of point defects, which are triggered by both mechanical and thermal stress. However, in organic and perovskite solar cells, the alternating combination of organic and inorganic interface materials are more and more being established as a design criterion due to the advantage of orthogonal solubility in multi-layer processing. Tensile strain then arises at the interface of materials with differing thermal expansion coefficients during temperature fluctuations. Such fluctuations dominantly occur during processing, annealing, and post-crystallization. The mismatch in thermal expansion coefficients can cause macroscopic delamination or cracking at interfaces as well as microscopic formation of point defects such as halide voids.

Furthermore, the molecular flexibility of interface materials, particularly their torsional flexibility, significantly affects the mechanical properties of thin films, including the Young's modulus. Molecules with higher torsional flexibility can accommodate strain more effectively, enhancing the mechanical resilience of the interface.

Recent research underscores that molecular tensile strain management might be regarded as a more general proxy for interface design. Brabec and Friederich reported in Science 2024  torsional flexibility of HTL molecules, which was represented by the introduction of TPA units, as one of several design criteria for efficient perovskite solar cells \cite{wu2024inverse}. We note that the suggestion "tensile strain + molecular architecture" was generated in December 2023. Properties such as film formation and homogenous film quality benefited from that as well. In another currently ongoing study by Brabec {\it et al.} (in submission), we found that embedding plasticizing molecules dramatically increases the mechanical and operational stability of thin film solar cells.

In conclusion, the effective management of tensile strain through the selection of interface materials with appropriate thermal expansion coefficients and molecular flexibility may have been underestimated for the design of long-time stable thin film organic and hybrid solar cells.

\subsubsection*{Suggestion: "Multiphase structure" + "Selective laser melting"}
% Horst Hahn

The term microstructure refers to the complete set of internal structural features of a material. This includes the crystallographic arrangement of its ideal, defect-free regions, as well as all types of defects that may be present. These defects range from point defects - such as impurity atoms, vacancies, and interstitials - to line defects like dislocations, planar defects such as grain boundaries, and three-dimensional defects including pores or precipitates. A well-characterized microstructure, from the atomic to the macroscopic scale, enables the establishment of correlations between structural imperfections and the mechanical or functional properties of materials. This understanding is essential for guiding the selection and application of materials in engineering contexts.

One important microstructural feature in complex metallic alloys and ceramic materials is the presence of phases - regions within the material that exhibit uniform physical and chemical properties, such as crystal structure and composition. Most technical materials are multiphase systems, meaning they contain multiple such regions. The formation of these multiphase structures typically results from processing steps that involve heating and cooling cycles, and often plastic deformation. These steps aim to achieve thermodynamically stable or kinetically stabilized microstructures. Controlling the multiphase structure is critical, as it allows materials to exhibit a desirable combination of properties, including enhanced strength, toughness, corrosion resistance, wear resistance, and optimized electrical or magnetic behavior.

Selective Laser Melting (SLM) is an advanced additive manufacturing (3D printing) technique in which a high-powered laser selectively melts and fuses fine layers of metal powder to fabricate solid, complex parts directly from a digital design. As the part is built layer by layer from metal powder, the process minimizes material waste while producing components with high detail and mechanical integrity.

Due to the rapid heating and cooling cycles inherent in SLM, the resulting microstructures are often far from equilibrium, frequently leading to the formation of multiphase materials. For example, parts produced via SLM may contain mixtures of phases with differing crystal structures or local compositions that would not typically form in slower, conventional manufacturing processes. These multiphase structures can significantly affect the material's final properties - potentially enhancing strength, hardness, or corrosion resistance, but also possibly introducing internal residual stresses that could compromise long-term performance.

\subsubsection*{Suggestion: "Stress induced phase transformation" + "Hexagonal boron nitride"}
% Christoph Kirchlechner

The link between stress-induced phase transformation and hexagonal boron nitride (h-BN) is best explained in the context of enhancing the fracture toughness of ceramics by phase transformations. For instance, in zirconia, a tetragonal-to-monoclinic phase transformation results in a volume expansion in the vicinity of a crack, which locally results in compressive stress both, deflecting and stopping crack growth. This concept is applied to other metastable phases in hard coatings \cite{hohenwarter2016}. A second concept of reaching an enormous tensile strength with simultaneously high fracture toughness, which both typically exclude each other, is using local anisotropy. This was shown in pearlite wires, where multiple micro-cracks parallel to the loading direction were formed and enhanced the fracture toughness substantially \cite{kupczyk2023}.
There are at least two arguments why thinking of stress-induced phase transformations and h-BN is reasonable. (i) Cubic boron nitride (c-BN), having a Wurtzite crystal structure, is used as a hard coating because of its enormous hardness. In contrast, the hardness of h-BN is significantly lower, rendering it not a top candidate for hard coatings. On the contrary, h-BN offers a significant anisotropy in its mechanical properties and therefore might locally help to produce micro-cracks parallel to the loading direction, resulting in toughening. (ii) c-BN and h-BN have different densities, with c-BN exhibiting the higher one. I.e., upon compression, a certain propensity of forming c-BN exists. However, this phase transformation seems not to be trivial, and it remains unclear if the c-BN to h-BN phase transformation could be exploited from a materials engineering viewpoint. While first indications of improved mechanical properties by using a c-BN / h-BN composite do exist \cite{krishnaiah2025}, the role of phase transformations to the best of our knowledge remains unknown for this system and could be an interesting topic for future evaluation.

\subsubsection*{Suggestion: "In-plane polarization" + "Organic solar cell"}
% Alexander Colsmann

A number of material properties were discovered in perovskite photovoltaics that have not been considered with respect to photovoltaics in the past, but which may benefit the generation and extraction of photogenerated charge carriers and hence the power conversion efficiency of the devices. Among these properties, a ferroelectric in-plane polarization of methylammonium lead iodide (MAPbI3) was vividly discussed to help with the separation of electrons and holes as well as their conduction to the respective electrodes along ferroelectric domain walls \cite{rossi_ferroelectric_2018,rohm_ferroelectric_2017}. So far, the ferroelectric properties are genuine to MAPbI3 solar cells and some derivatives thereof \cite{rohm_ferroelectric_2019}. The necessary precondition is a non-centrosymmetric unit cell of the crystal, which is why this concept has not played a role in any of the established solar technologies, such as silicon, so far \cite{rohm_ferroelectric_2019,breternitz_role_2020}. Yet, it may well play a role in future novel perovskite-inspired materials \cite{krishnaiah2025}.

The requirement of crystallinity and non-centrosymmetric unit cells seemingly does not link to organic solar cells, which are mostly amorphous. Yet, the great design flexibility of organic molecules can be used to introduce asymmetric polar functional groups, which are already in use today in piezoelectric polymers for organic touch sensors, e.g., polyvinylidene fluoride (PVDF). If, in addition, ways are found to orient these molecules, an organic analogy to in-plane polarization may be designed, which then also benefits the charge carrier separation in organic solar cells. The challenge here is that in-plane polarization is notoriously difficult to control via external electrical poling since the electrodes in organic solar cells only allow for the application of an out-of-plane electric field.~\cite{roehm2020} Hence, in-plane polarization probably must be part of the conformation of the film itself through post-processing or by molecular self-assembly. Covalent organic frameworks (COFs) may also mimic a crystal structure, which then can be modified for in-plane polarization. 

Furthermore, polarizable molecules typically have higher dielectric constants, which reduce the binding energy of photo-generated electron-hole pairs. While this binding energy still limits the performance of organic solar cells today, increasing the dielectric constant through polarization may help to enhance the open-circuit voltage of solar cells and to reduce monomolecular recombination. The idea of increasing the dielectric constant of organic molecules to reduce the binding energy of electron-hole pairs has been around for many years, yet without any major breakthroughs. The use of polarizing functional groups has not been thoroughly investigated so far.

As of today, both resulting concepts are rather speculative, but the machine learning suggestion points in an interesting research direction to be considered in future experiments.

\section{Dataset}\label{si-sec:dataset}
%\noindent
\paragraph{Filtering}
Papers fetched from venues related to materials sciences %\ref{table:excerpt-sources}
were first filtered, and only 259,913 out of 490,188 works had both a title and an abstract. Additional filtering removes very short (< 100 characters) and very long (> 4000 characters) abstracts. Furthermore, we check relatedness to the materials science domain by checking whether the tag 'Materials Science' is present. We rely on tags from OpenAlex's \href{https://github.com/ourresearch/openalex-concept-tagging}{tagger}. We exclude works that contain Latex code and those whose primary language is not English.

\paragraph{Cleaning} 
We exclude non-ASCII characters, remove tabs, multiple whitespaces and newlines. We use heuristic approaches to replace latex super or subscript code. We replace number ranges, parenthesis and brackets.

\begin{sifigure}[t!]\label{si-fig:data-fetching-cleaning}
    \centering
    \includegraphics[scale=0.2]{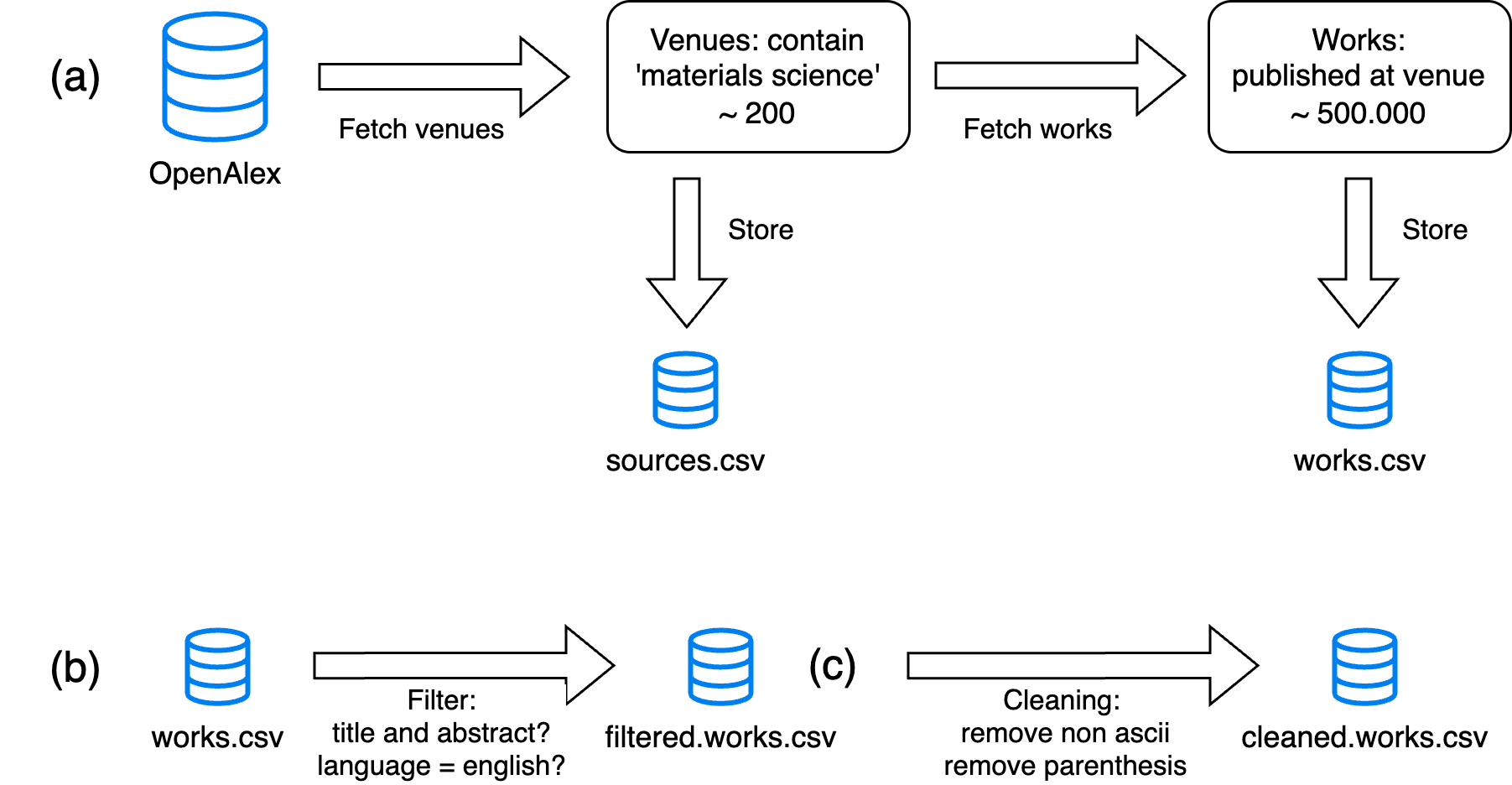}
    \caption{Overview of the dataset generation: (a) fetch publications from \href{https://docs.openalex.org/}{OpenAlex} venues related to materials sciences,  (b) filtering based on title and abstract (c) removal of selected characters.}
\end{sifigure}

\paragraph{Chemical Formulas} Predominantly found in the materials science domain are material formulas. We employ a custom parser to recursively identify materials following these five rules:
\begin{enumerate}
    \item The subsequent word is another element
    \item The subsequent word is an integer together with a + or -
    \item The subsequent word is an integer, a + or a - 
    \item The subsequent word is a decimal
    \item The subsequent word is a composition of elements of the periodic table
\end{enumerate}

The following examples showcase how those rules translate into real parsing:
'Na Cl' $\rightarrow$ 'NaCl',
'Na 2' $\rightarrow$ 'Na2',
'Mn 4+' $\rightarrow$ 'Mn4+',
'Fe3 O' $\rightarrow$ 'Fe3O',
'Fe3O 4' $\rightarrow$ 'Fe3O4',
'Mn 0.75' $\rightarrow$ 'Mn0.75',
'Zn3 PO' $\rightarrow$ 'Zn3PO', and
'H2 SO4' $\rightarrow$ 'H2S04'.

\section{LLM Fine-tuning}\label{si-sec:fine-tuning}
Hugging Face's Python library was used to fine-tune the LLaMa-2-13B model \cite{touvron_llama_2023-1, wolf2020}. We prepare our annotated data by constructing a single string per data point as follows \texttt{<s> + abstract + "\#KEYWORDS\#" + concepts + </s>} where \texttt{concepts} corresponds to the string version of a Python list, e.g. \texttt{['brittle epoxy', 'butadiene nitrile liquid rubber', ...]}. \texttt{<s>} and \texttt{</s>} represent the beginning of sequence (BOS) and, respectively, end of sequence (EOS) token for the LLaMa models. \texttt{\#KEYWORDS\#} serves as an own custom delimiter to signal the model when concept extraction should start. For the LLaMa tokenizer, we set \texttt{pad\_token\_id} = 0, \texttt{bos\_token\_id = 1}, and \texttt{eos\_token\_id = 2}. Being a CausalLM, we set the labels equal to the input token ids. Our \lstinline{LoraConfig} is \lstinline{r = 16} (Lora attention dimension), \lstinline{lora_alpha = 32}, \lstinline{target_modules=["q_proj", "v_proj"]} (modules where the adapter is applied), and \lstinline{lora_dropout = 0.01} (dropout probability) \cite{dettmers2023qlora}. Training took around an hour on a single A100 GPU.

\section{Hyperparameters of the link prediction models}\label{si-sec:hyperparameters}
\begin{sitable}[ht]
    \centering\small
    \caption{Optimized hyperparameters of link prediction models.\textsuperscript{[a]}}
    \begin{tabular}{p{0.3\linewidth}p{0.1\linewidth}p{0.35\linewidth}p{0.15\linewidth}}
        \toprule
        Model & Layers & Neurons per layer & Learning rate \\ \midrule 
         Baseline & 7 &  [20, 300, 180, 108, 64, 10, 1] & $5 \cdot 10^{-4}$\\
         Concept embeddings & 5 & [1536, 1024, 819, 10, 1] & $1 \cdot 10^{-3}$\\
         Combination of features & 7 & [1556, 1556, 933, 559, 335, 10, 1] & $1 \cdot 10^{-3}$\\ \bottomrule
         \addlinespace[\belowrulesep]
         \multicolumn{4}{p{0.9\linewidth}}{[a] A ReLU activation function, a dropout probability of $p_{\text{dropout}} = 0.1$ and 30\% of positive samples per batch were used for all models.}
    \end{tabular}
    \label{si-tab:hyperparameters}
\end{sitable}

The GNN link prediction model uses a two-layer GraphSAGE encoder with a hidden dimension of 256 and produces 128-dimensional node embeddings. These embeddings are then fed into an MLP decoder that also has a hidden dimension of 256 and operates on 128-dimensional inputs. Neighborhood sampling is performed with fanout $(k_1,k_2) = (20,15)$ for the first and second hop, respectively. The model is trained with a learning rate of $1 \cdot 10^{-5}$; additionally, a ReLU activation function, dropout with $p_{\text{dropout}} = 0.1$ is used.

\emergencystretch=2.5em

\end{document}